\documentclass{article}
\usepackage[nonatbib]{poolside_report}

\usepackage[utf8]{inputenc}    %
\usepackage[
  backend=biber,
  style=numeric-comp,   %
  natbib=true,
  url=false,
  doi=false,
  eprint=false,
  isbn=false,
  giveninits=true,
  maxbibnames=8,
  minbibnames=3,
  backref=true,
]{biblatex}
\usepackage[T1]{fontenc}       %
\usepackage{hyperref}          %
\usepackage{url}               %
\usepackage{xurl}
\AtBeginBibliography{%
  \raggedright
  \emergencystretch=2em
  \hyphenpenalty=50
  \tolerance=400          %
}
\usepackage{booktabs}          %
\usepackage{amsfonts}          %
\usepackage{amsmath}
\usepackage{amssymb}
\usepackage{nicefrac}          %
\usepackage{microtype}         %
\usepackage{xcolor}            %
\usepackage{graphicx}          %
\usepackage{subcaption}        %
\usepackage{multirow}          %
\usepackage{algorithm}         %
\usepackage{algpseudocode}
\usepackage{placeins}          %
\usepackage{float}             %
\usepackage{cleveref}          %
\usepackage{tikz}              %
\usetikzlibrary{arrows.meta}
\usepackage[normalem]{ulem}              %
 
\hypersetup{
    colorlinks=true,
    linkcolor=blue!50!black,
    citecolor=blue!50!black,
    urlcolor=blue!50!black,
}

\AtEveryBibitem{
\clearfield{abstract}
\clearfield{pages}
}
\newcommand{\laguna}{\textsc{Laguna }}
\newcommand{\lagunam}{\textsc{Laguna~M.1 }}
\newcommand{\lagunamshort}{\textsc{M.1 }}
\newcommand{\lagunaxs}{\textsc{Laguna~XS.2 }}
\newcommand{\lagunaxsshort}{\textsc{XS.2 }}

\title{Laguna M.1/XS.2 Technical Report}

\author{%
  Poolside team\\
  \texttt{research@poolside.ai} \\\\
  \url{https://huggingface.co/poolside/}
}

\begin{document}

\maketitle

\pagestyle{plain}

\begin{abstract}
We present \lagunam and \lagunaxs\!\!, two Mixture-of-Experts foundation models built for long-horizon, agentic coding: \lagunamshort has $225.8$B total parameters ($23.4$B activated per token) and \lagunaxsshort has $33.4$B total ($3$B activated). Both models were trained from scratch end-to-end inside the same internal system that we refer to as our \emph{Model Factory}: a tightly-integrated stack of versioned data, training, evaluation, and inference components that turn model development into an industrial process. We describe the principles and design choices of the Model Factory and also detail the end-to-end training process of our models, throughout pre-training data and architecture, post-training stages, evaluation, and quantization.

On agentic software engineering and terminal benchmarks (SWE-bench Verified, SWE-bench Multilingual, SWE-Bench Pro, and Terminal-Bench 2.0) \lagunamshort and \lagunaxsshort are competitive with state-of-the-art open models in their respective weight classes. \lagunaxs weights are released under Apache~2.0 at \url{https://huggingface.co/collections/poolside/laguna-xs2}.
\end{abstract}

\newpage

\section{Introduction}

In this report, we present \lagunam and \lagunaxs\!\!, two foundation models built in quick succession, together with the engineering-first \emph{Model Factory} methodology we used to build them. \lagunamshort and \lagunaxsshort are capable agentic coding models built for long-horizon, multi-step work, competitive with state-of-the-art models in their weight class. We believe mastering agentic coding is the key to unlocking generalized autonomous problem solving for all knowledge tasks, since code is the most general medium for expressing and verifying solutions to structured problems.

\lagunamshort and \lagunaxsshort are Mixture-of-Experts foundation models. \lagunamshort has $225.8$B total parameters with $23.4$B activated per token; \lagunaxsshort has $33.4$B total with $3$B activated. Both models were trained from scratch, end-to-end, inside the same internal system, which we refer to as the \emph{Model Factory}. One focus of this report is the construction process itself, illustrated by the lessons we learned building \lagunamshort\!\!, and implementing them in quick succession to build \lagunaxsshort afterwards. We started \lagunaxsshort just after pre-training of \lagunamshort finished, and the end-to-end time from start of training to release of \lagunaxsshort spanned only five weeks. The weights of \lagunaxsshort are available to the public under the Apache~2.0 license, and can be accessed via \url{https://huggingface.co/collections/poolside/laguna-xs2}.

\begin{figure}[h]
  \centering
  \makebox[\textwidth][c]{%
  \begin{minipage}{\textwidth}
    \centering
    \begin{subfigure}[t]{\textwidth}
      \centering
      \includegraphics[width=\textwidth]{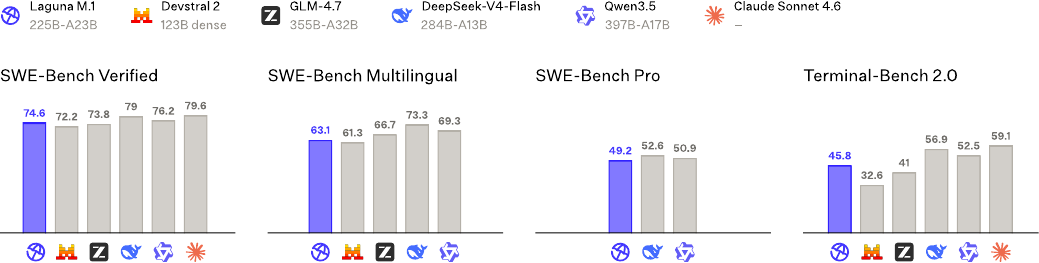}
      \caption{\lagunamshort results.}
      \label{fig:lagunam_results}
    \end{subfigure}

    \par\addvspace{24pt}

    \begin{subfigure}[t]{\textwidth}
      \centering
      \includegraphics[width=\textwidth]{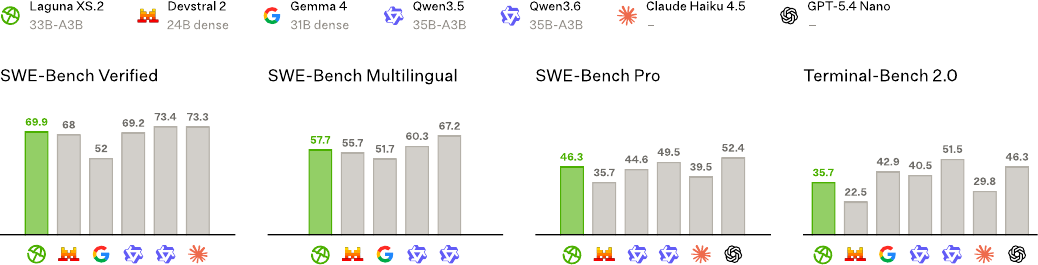}
      \caption{\lagunaxsshort results.}
      \label{fig:lagunaxs_results}
    \end{subfigure}
    \vspace{5pt}
    \caption{\lagunam and \lagunaxsshort results on agentic benchmarks compared to Devstral 2 and Devstral Small 2~\citep{mistral2025devstral-2}; Qwen3.5 397B-A17B and 35B-A3B~\citep{qwen35blog}; Qwen3.6 35B-A3B~\citep{qwen36_35b_a3b}; GLM-4.7~\citep{glm47team2026}; DeepSeek-V4 Flash~\citep{deepseek_v4}; Claude Sonnet 4.6~\citep{anthropic2025claudesonnet46} and Haiku 4.5~\citep{anthropic2025claudehaiku45}; Gemma 4 31B dense~\citep{gemma4}; and GPT-5.4 Nano~\citep{openai2026gpt54nano}.}
    \label{fig:results}
  \end{minipage}%
  }
\end{figure}

\paragraph{Results.}
Both models are competitive with state-of-the-art models of comparable size on agentic tasks. \Cref{fig:results} shows our evaluation results on agentic software engineering and terminal task benchmarks: SWE-bench~Verified, SWE-bench~Multilingual, SWE-Bench~Pro, and Terminal-Bench~2.0. Details of our evaluation protocol can be found in \Cref{sec:evaluation}.

\paragraph{Model Factory Approach.}
In addition to sharing technical details of \lagunam and \lagunaxsshort\!\!, we want to draw attention to the underlying process that produced these models. 
Building \lagunaxsshort from inception to delivery in five weeks was only possible because we treat foundation model development as an industrial process. We have built a set of components and processes which we call our \emph{Model Factory}, with the aim to accelerate the process of research and model building, so that our time can be focused on defining and investigating genuine research questions, while integration and plumbing are automated as much as possible. We describe the guiding principles of our Model Factory in \Cref{sec:factory}, and show examples of how this works in practice throughout this report.

\paragraph{Technical Overview.}
We built \lagunamshort and \lagunaxsshort from scratch, and share technical details on our Model Factory, architecture, pre-training data, post-training, quantization, and evaluation pipeline, in the hope of contributing to open research.

\begin{itemize}
  \item \textbf{Model Factory.} We describe the design philosophy of our Model Factory, and illustrate the principles that we follow to make model building and research into a repeatable industrial process rather than artisanal projects in \Cref{sec:factory}.
  \item \textbf{Pre-training.} We share details of our pre-training in \Cref{sec:pre-training}. This includes architecture design choices for both models, findings on efficiency and stability when training large MoEs from scratch.

  In addition, we also go into our pre-training data strategy. Both \laguna models were trained across more than 30T tokens during pre-training, drawn from a mix of web, code, and synthetic sources. We describe our data curation and preparation pipeline, as well as our data mixture tuning.
  \item \textbf{Post-training.} We describe our post-training pipeline in \Cref{sec:post-training}, which is divided into imitation learning stages and a final reinforcement learning stage. We discuss how the recipe transferred between \lagunamshort and \lagunaxsshort\!\!, and the data and formatting choices that proved critical for stable training.
  \item \textbf{Quantization.} To make \lagunaxsshort deployable on low-VRAM devices, we quantize the model's MoE layers to \texttt{FP8}, \texttt{INT4}, and \texttt{NVFP4}, as well as the KV cache to \texttt{FP8}. We describe these quantization recipes and the mixed-precision strategy we adopted to preserve quality in \Cref{sec:quantization}.
  \item \textbf{Evaluation.} We describe our benchmark selection and evaluation harness, and present results on base models, as well as agentic software engineering and terminal task benchmarks in \Cref{sec:evaluation}.
\end{itemize}

\section{Model Factory Approach}\label{sec:factory}
We invest heavily into continuously improving the \emph{process} of how we conduct research and model building, even moreso than into any individual model. This has resulted in a process that we internally call our \emph{Model Factory}: a set of components and processes with the aim to maximize research iteration and integration speed, and to minimize the amount of attention researchers need to pay to bookkeeping and infrastructure. This process that allowed us to build \lagunaxs from scratch to delivery within five weeks, applying the lessons learned from training \lagunamshort\!\!.

\subsection{Factory Principles}
Running research and model building at scale comes with a few challenges: how do we coordinate across many independent research streams; how do we continue to improve research velocity; how do we quickly integrate research results into big production runs? 

We share our principles to address these questions in the following sections. Throughout the report, we also show examples of these play out in practice in the respective technical sections.

\subsubsection{Experiments as Code}
All inputs and configurations to any run (data pipelines, ablation experiments, any training run) are expressed as code committed into a single repository. Each run is tracked with a unique ID, providing a stable handle for identifying, tracking, and reasoning about lineage across experiments and runs. Correspondingly, all experiments, their inputs, and artifacts are registered as persistently stored assets that track dependencies to each other. This gives us the following benefits.

\paragraph{Control Plane.}
We use Dagster~\citep{dagster} as the central control plane to navigate the directed acyclic graph (DAG) of assets, giving us answers to two questions: \emph{what runs}, and \emph{what does each run depend on}. It also provides a single entrypoint for researchers to manage their experiments: Any job is launched by registering a configuration asset, and kicked off either via CLI or UI. Keeping track of runs allows us to maintain a full reproducible history of what experiments were run, with what configuration and inputs, and where to find results and outputs.

\paragraph{End-to-end Lineage.}
Maintaining a DAG of inputs, runs, and artifacts allows us to surface the complete lineage of any model or artifact in both directions: a token in a packed pre-training shard can be traced back through deduplication, filtering, and synthesis to its source document. Every checkpoint, evaluation result, and inference deployment can be traced back to the training run that produced it.  This property is a force multiplier to our researchers, giving them full visibility into the combined research results of the whole team, enabling them to branch off any interesting idea they find without needing to manually piece together its lineage.

\paragraph{Agentic Workflows.}
Our control plane also allows AI agents a single entry point to participate in the research process. Through it, and our code base, they can access the full history of existing research, and corresponding assets and artifacts. Today we use agents to design and run ablation experiments, monitor runs and debug issues, and compile and analyze experiment results. Over time we expect them to take an even greater role in day-to-day research, autonomously pursuing and validating research directions.

\subsubsection{Composable, Decoupled Components}
We pursue the ideal that every part of the model pipeline should be built once, to be reused and composed many times in different places. This is not an obvious choice as often research and production live in different code bases, driven by different needs for flexibility.

We make the point that maintaining a single code base for research and production (including different research streams) gives us a crucial advantage in making any research progress immediately available to everyone in the team, and widely to users in production. This requires a continued discipline in maintaining a code base that can serve different purposes.

These components cover every step in our research and model pipeline, including 
data pipelines, training, reinforcement learning, inference, and evaluation. The composability allows us to iterate on components in isolation while making sure the end-to-end flow works, such that a successful innovation can be promoted into production by simply flipping a configuration flag. We highlight some examples of our components in the remaining sections, where they are used across pre-training, post-training, inference, and evaluations.

\subsubsection{Reserve Human Attention for Novel Decisions}
Another core principle of our Model Factory is to minimize the need for repetitive and mechanical manual work as much as possible, focusing researcher's attention on genuine research questions. 

One of the key pieces that enables this is our custom scheduling system that allows researchers to deploy and run training and inference workloads with minimal need to care about the underlying orchestration and resource management. Great care is also spent on automated training failure recovery, and incident escalation all run without engineer involvement on the happy path. On-call is paged only when the automatic recovery system fails to make progress.

\paragraph{Custom Workload Scheduler.}
Jobs are scheduled by Kubernetes~\citep{kubernetes} across multiple node pool types. Topologies range from single-node debugging runs to full-cluster jobs on the order of $10^4$ accelerators.

The cluster is managed by our custom cluster scheduler. An initial version was based on Volcano~\citep{volcano}. Over time two limitations became binding: First, Volcano makes eviction and reclaim decisions per node. When capacity needs to be freed for a higher-priority gang, entire nodes are drained, which displaces unrelated co-tenant pods and forces them to re-queue from scratch. Second, the scheduler relies on the Kubernetes API server, and therefore on etcd, to hold cluster topology. Under sustained churn at our scale, etcd latency dominated scheduling decisions and pushed end-to-end placement times into tens of minutes.

We replaced it with an in-house batch scheduler purpose-built around three observations:

\begin{enumerate}
  \item \textbf{Per-job eviction and reclaim.} Capacity decisions are made at the granularity of jobs, not nodes. When a higher-priority job needs resources, the scheduler selects victim jobs whose preemption minimizes total displaced work, rather than draining whichever nodes happen to host the required slots. Co-tenant jobs whose pods are not directly required for reclaim continue running undisturbed.
  \item \textbf{Topology in FoundationDB, populated by observers.} Cluster topology --- nodes, pods, GPUs, NVLink and fabric groupings, taints, and capacity --- is mirrored out of Kubernetes into FoundationDB~\citep{foundationdb} through an observer-based controller that watches the API server and reconciles changes incrementally. The scheduler reads from FoundationDB, which sustains the read and transactional write rates we need without the latency tail we observed when scheduling decisions were issued directly against etcd.
  \item \textbf{Sticky pod respawn.} When a pod within a running job dies --- whether from a node-level fault, a transient NCCL failure, or a preemption that is later reversed --- the scheduler prefers to respawn it on the same node it was previously bound to, provided the node is healthy. This dramatically reduces churn: caches stay warm, fabric topology assumptions remain valid, and the surrounding gang does not have to re-pack across the cluster.
\end{enumerate}

The cumulative effect on tail latency was the change that mattered most operationally. With our custom scheduler, placement is consistently sub-minute. This made the rest of the factory's automation viable at our scale: hyperparameter sweeps, CI canaries, and preemption-driven backfill all rely on the scheduler being predictably fast.

\section{Pre-training}\label{sec:pre-training}

In this section, we describe the details of our architecture and training setup, the construction of our pre-training data, and present evaluations of our base model on pre-training benchmarks. We outline the training instabilities we encountered with \lagunam\!\!, the lessons drawn from them, and how those lessons shaped \lagunaxs\!\!. We report evaluation results in \Cref{sec:evaluation}.

\subsection{Architecture \& Training}\label{sec:architecture}
\lagunaxs and \lagunamshort are both Mixture-of-Experts (MoE)~\citep{shazeer2017moe} models using a pre-norm Transformer architecture~\citep{vaswani2017transformer, xiong2020prenorm} with RMSNorm~\citep{zhang2019rmsnorm} as the normalization layer. \lagunaxs has a total parameter count of 33.4B with 3B parameters (including embeddings) activated per token, while \lagunam has 225.8B total parameters out of which 23.4B (including embeddings) are activated per token. \lagunaxs uses token-choice routing~\citep{shazeer2017moe} with 8 of 256 experts activated per token, plus a shared expert~\citep{dai-etal-2024-deepseekmoe} that processes every token. The output of the routed experts is modulated by a coefficient of $2.5$ before combining with the output of the shared expert. We use a linear layer followed by a sigmoid activation as the routing function, and apply normalization to the scores after top-k~\citep{liu2024deepseek}. For load-balancing, we use an auxiliary loss from \citet{qiu-etal-2025-demons} across the non-padding tokens of the global batch. For training stability, the first Transformer layer is dense. Both models share the same tokenizer with a vocabulary size of $100{,}352$ tokens, trained with Byte-Pair Encoding~\citep{sennrich-etal-2016-neural} on our internal datasets.

\lagunaxs uses interleaved Sliding Window Attention (SWA) and Global Attention (GA)~\citep{beltagy2020longformer, gemmateam2024gemma2improvingopen} with a 3:1 ratio. In both types of layers, we use Grouped Query Attention (GQA)~\citep{ainslie-etal-2023-gqa} with 8 KV heads, a head dimension of $128$, and softplus-based per-head gating~\citep{qiu2026gated}. We employ Rotary Positional Encodings (RoPE)~\citep{su2024rope} to encode positional information. GA layers use 48 Q-heads, $\theta = 500{,}000$, and partial RoPE applied to the first 50\% of the head dimension~\citep{wang2021gptj}. In layers using SWA, we use an attention window of $512$ tokens, 64 Q-heads, and a smaller $\theta = 10{,}000$. The architecture choices above were selected via ablations on a smaller MoE proxy; ablation results are available in \Cref{app:swa_ablations}.

\paragraph{From \lagunam to \lagunaxs\!\!.} After training \lagunam\!\!, we revisited many of the architecture and data decisions, including some that had been made implicitly rather than deliberately. On the architecture and training side, \lagunaxs inherited many well-ablated choices from \lagunam but introduced four notable changes. First, the more efficient attention mechanism described above replaces global attention on every layer in \lagunam\!\!. Second, we adopted a Warmup-Stable-Decay (WSD) learning rate schedule~\citep{hu2024minicpm} instead of cosine. Third, we added routed expert modulation, similar to DeepSeek-V3~\citep{liu2024deepseek} and Nemotron 3~\citep{nemotronnano}. Fourth, we reduced the number of dense layers at the bottom of the model from 3 to 1. We paid particular attention to hyperparameters and numerical stability after the \lagunam training instabilities described in \Cref{sec:training_stability_grad}, and established a scaling law for learning rate prediction (\Cref{sec:pretraining_lr_law}). On the data side, we added an AutoMixer (\Cref{sec:automixing}), substantially reduced repetitions across smaller high-quality datasets through synthetic rephrasing (\Cref{sec:synth_data}), and increased diversity of our web data pipeline by reworking our filters.

Revisiting and adopting these decisions for \lagunaxsshort was cheap precisely because of our Model Factory approach (\Cref{sec:factory}): the work on data pipelines, training stack, ablation proxy, and evaluation harness transfers seamlessly to any new training run, and the questions that remained were purely model-design questions. Each of the four deltas was selected by a series of targeted ablations on our smaller MoE proxy, and promoted to \lagunaxs as a configuration change against the \lagunam baseline.

\subsubsection{Training Recipe}\label{sec:training_recipe}

The training recipe for \lagunaxs reflects the changes outlined above (\Cref{sec:architecture}): a WSD schedule rather than cosine, peak LR set by the scaling law in \Cref{sec:pretraining_lr_law}, and numerical conventions informed by the stability investigations in \Cref{sec:training_stability_grad}. \lagunam and \lagunaxs were pre-trained on $6{,}144$ and $2{,}048$ NVIDIA H200 GPUs, respectively. We use the Muon optimizer~\citep{jordan2024muon}, specifically the Moonlight variant of \citet{liu2025muon}, across all training stages, including SFT and RL --- \Cref{sec:muon} provides details on our distributed Muon implementation.

We use \texttt{BF16} mixed precision training throughout all stages with the master weights in \texttt{FP32} and most computations in \texttt{BF16}. Exceptions include RMSNorm layers and RoPE, which use selective \texttt{FP32} operations, and the LM head input-gradient all-reduce described in \Cref{sec:training_stability_grad}. The pre-training stage uses a context length of 4K tokens and a WSD schedule: linear warmup to a peak learning rate of $5{\times}10^{-4}$, a stable phase at peak, and a final decay phase covering the last $30\%$ of training steps that follows a $1{-}\sqrt{\cdot}$ shape and ends at $5\%$ of the peak ($2.5{\times}10^{-5}$).

\paragraph{Peak Learning Rate from a WSD Scaling Law.}\phantomsection\label{sec:pretraining_lr_law}
WSD is attractive because a single stable-phase checkpoint can be reused
across data-mix iterations via shorter cooldown runs. But, according to our ablations, final-loss
calibration is less direct than for one-shot cosine, and naively tuned
WSD runs sometimes underperform tuned cosine at matched compute. To pick
the peak LR for \lagunaxs at production scale, we therefore fit a WSD-specific
scaling law by sweeping four MoE sizes (2B--16B
total / 0.3B--2.2B active) across six learning rates and five token
budgets (30B--480B) at fixed batch $B_0 = 8$M tokens on our stage-1 mix.
For each $(N, D)$ we extracted the optimum $\text{lr}^\star(N, D)$ from
a parabola fit in $\log_{10}\text{lr}$ space, then used ordinary least squares regression to fit a global power law (full procedure, per-cell plots, and robustness checks in
\Cref{app:lr_scaling}):
\begin{equation}
  \text{lr}^\star(N, D) \;=\; 10^{4.488} \cdot N^{-0.4639} \cdot D^{-0.2661},
  \label{eq:lr_law}
\end{equation}
with $N$ the number of active parameters and $D$ the total token budget
(cooldown included). For a different global batch $B$ we scale the
predicted LR by $\sqrt{B/B_0}$, following standard square-root batch--LR
scaling for Adam-family optimizers~\citep{malladi2024sdes}. \lagunaxs
has $N = 3.0$B active parameters and trains at $B = 24$M tokens; the
law predicts $\sim 5.5{\times}10^{-4}$, and we use $5{\times}10^{-4}$ to leave
a small safety margin.

\paragraph{Long-Context Training.} We start the context extension from the end-of-decay checkpoint and split it into two equal-token sub-stages of $100$B tokens each. The first sub-stage extends the context to $32$K tokens; the second extends the context to $128$K tokens. Both sub-stages apply YaRN~\citep{peng2024yarn} to global attention layers only, share a global batch size of $24$M tokens, and use a cosine schedule that decays from $5\%$ of the pre-training peak ($2.5{\times}10^{-5}$) to $1\%$ of the pre-training peak ($5{\times}10^{-6}$). We do not re-warm up the learning rate and resume directly from the end-of-decay pre-training checkpoint; in our experiments this transferred better than reintroducing a warmup at the start of context extension. To reduce variation in the final base checkpoint, we apply an exponential moving average (EMA) over the 10 most recent checkpoints from the end of the $128$K stage and use the EMA weights as the final base checkpoint. For the final checkpoint, we extend the context length further to $256$K without any training by doubling the RoPE scale in the global attention layers.

\subsubsection{Distributed Training}\label{sec:distributed_training}

\begin{notebox}[Model Factory Component: Titan]
  Titan is a PyTorch-based~\citep{pytorch} training library, originally seeded from TorchTitan~\citep{liang2025torchtitan} and extensively adapted (more than 2{,}200 changes) for our architecture, sharding strategies, checkpointing, and observability needs. It serves as the single training entry point for the factory, and handles pre-training and post-training needs, including reinforcement learning.
\smallskip
It supports composing the full set of
standard sharding paradigms --- Distributed Data Parallel (DDP), Fully
Sharded Data Parallel (FSDP)~\citep{zhao2023fsdp}, Tensor Parallel (TP)
\citep{shoeybi2019megatron}, Expert Parallel (EP)~\citep{shazeer2017moe}, and Pipeline Parallel (PP)~\citep{huang2019gpipe} --- with careful overlap of communication
and computation. The training loop, data loader, and metrics path are
aggressively optimized for low CPU overhead so that the GPU runs
consistently ahead. We have also paid close attention to numerical
correctness throughout the codebase; some representative examples include
resetting position indices at document boundaries to avoid RoPE-related rounding
under document masking in attention, and making gradient-reduction dtypes
configurable per collective with \texttt{FP32} as the default wherever a reduction is numerically sensitive (see \Cref{sec:training_stability_grad}).
\end{notebox}

\paragraph{Device Mesh.} A full model/optimizer replica is distributed
across FSDP, the TP/EP dimensions, and PP, with replication across DDP
ranks. The mesh dimensions are
$(\mathrm{PP},\mathrm{DDP},\mathrm{FSDP},\mathrm{TP})$ for non-MoE layers,
and $(\mathrm{PP},\mathrm{DDP},\mathrm{FSDP},\mathrm{EGP},\mathrm{ETP})$ for
MoE layers. Here, Expert Group Parallel (EGP) shards experts across ranks
(analogous to standard EP) and Expert Tensor Parallel (ETP) shards the
weights of a single expert along its tensor dimension. Together, EGP and ETP take the place of TP in MoE layers, with
$\mathrm{EGP}\times\mathrm{ETP}=\mathrm{TP}$. For \lagunaxs and \lagunam we
set $\mathrm{ETP}=1$, since the per-expert intermediate dimension is small
enough that further tensor-parallel sharding does not pay off.

\paragraph{Tensor- and Sequence-parallel Non-MoE Components.} Each
component that lives on the TP dimension in a non-MoE layer (attention,
normalizations, embeddings, LM head) can be configured individually as
either tensor-parallel (TP) or sequence-parallel (SP).
For most components SP shards activations along the sequence dimension
in the usual sense~\citep{korthikanti2023sequence}; for attention,
however, we shard along the batch dimension in order to keep individual
sequences intact, so that attention is computed without cross-rank
communication. What we call sequence-parallel
attention is sometimes referred to as data-parallel attention in the
literature. When a component is switched from TP to SP, its weights are
no longer sharded along the TP dimension and are instead replicated
across TP ranks; each SP rank then produces only a partial weight
gradient, computed on its local slice of the input stream, and these
partial gradients must be summed across TP ranks before being passed to
the FSDP/DDP gradient reductions. We perform these TP weight-gradient
all-reduces asynchronously with respect to the compute stream, by
carefully adapting PyTorch's FSDP post-backward logic so that the
all-reduce is queued on a dedicated communication stream and overlapped
with subsequent backward computation.

\paragraph{Distributed Muon.}
\phantomsection\label{sec:muon}

As noted above, we use the Muon optimizer throughout. Compared to many other optimizers, Muon incurs a significant computational overhead that we tackle through the distribution of the compute across ranks. At a high level, Muon needs to aggregate the gradients into a momentum buffer, apply Nesterov momentum, orthogonalize them via Newton--Schulz, and update the parameters. Naively, each rank would need to do this for every full parameter. Our implementation assigns each parameter and gradient to only one of the ranks sharding it, gathers the full gradient on that rank, performs Newton--Schulz, and redistributes the corresponding orthogonalized gradient shards back to all other ranks within the group, which then update their local parameter shards. That effectively removes the compute bottleneck of the Muon optimizer, at the cost of additional communication. Following \citet{amsel2026the}, we use a schedule of coefficients for the Newton--Schulz iterations rather than reusing the same coefficients across iterations.

Our implementation overlaps batched communication with the Newton--Schulz computations. We also support enabling CUDA graphs for the Newton--Schulz procedure to reduce the CPU overhead of launching many relatively small kernels; this is mainly beneficial for smaller models. Combined, these optimizations reduce the optimizer overhead to less than 1\% of the training step time during \lagunam pre-training.

\paragraph{Compute-communication Overlap for MoE.} For MoE layers, we use Expert Parallelism, sharding expert weights across 8 H200 GPUs interconnected via NVLink. This requires two all-to-all collectives per layer: a \emph{dispatch} to send tokens to the GPUs owning their expert weights, and a \emph{combine} to retrieve the processed tokens back. Inspired by ParallelKittens~\citep{sul2025parallelkittenssystematicpracticalsimplification}, we fuse the \emph{dispatch} and \emph{combine} directly into the standard CUTLASS-based grouped GEMM kernels used by PyTorch, minimizing the idle time of Tensor Cores.

\begin{figure}[h]
  \centering
  \includegraphics[width=1.15\textwidth]{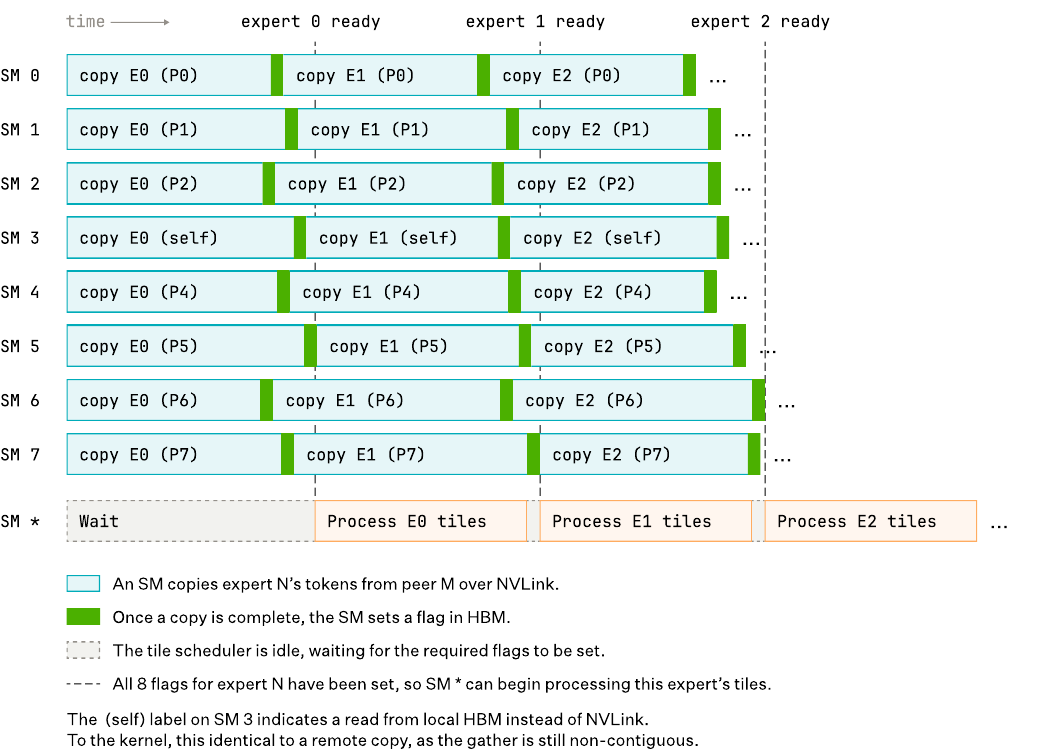}
  \vspace{5pt}
  \caption{Our strategy for overlapping the \emph{dispatch} with the grouped GEMM compute. Time flows from left to right. SMs numbered from 0 to 7 copy tokens. SM* illustrates all SMs running the grouped GEMM.}
  \label{fig:dispatch_overlap}
\end{figure}

For \emph{dispatch}, we dedicate 8 out of 132 Streaming Multiprocessors (SMs) on each H200 GPU to copying expert tokens from its peer GPUs to local High Bandwidth Memory (HBM). The tokens are copied in increasing expert order via vectorized 128-bit loads over NVLink, and a flag in the HBM is set once all tokens for a given expert have arrived. The remaining SMs run the original grouped GEMM algorithm, with the tile scheduler modified to wait until all expert flags for a tile are set, ensuring the tile has been fully copied before processing (see \Cref{fig:dispatch_overlap}).

For \emph{combine}, we modify the grouped GEMM epilogue that normally stores an output tile from the SM registers to the HBM. The modified epilogue instead rearranges the results in a layout amenable to vectorized 128-bit stores through shared memory, then sends each processed token directly to its owner over NVLink. Since epilogues have no data dependencies on one another, all SMs can run in parallel.

This scheme overlaps grouped GEMM compute and NVLink communication at a fine granularity, and is straightforward to integrate into any grouped GEMM kernel that exposes a tile scheduler and an epilogue.

Additionally, we reserve 5 SMs on every GPU for NCCL kernels used in data-parallel communication. Specifically, 4 SMs are used for AllGather/ReduceScatter FSDP collectives, and 1 SM is used to aggregate auxiliary loss information. \emph{Dispatch}/\emph{combine} collectives utilize the scale-up network, while data-parallel communications use the scale-out network; hence, they do not contend for bandwidth.

\begin{notebox}[Model Factory: Reliability]
Hardware faults at our cluster scale range from slowdowns to silent corruption of an entire training run. Pre-flight checks stress-test every node before admission; an in-flight recovery system detects hangs, NCCL failures, and node losses, replaces affected workers, and resumes from the latest checkpoint without human involvement. Critically, we found hash checks important to prevent silent and non-silent failures of large-scale training runs.

\paragraph{Cross-replica Hash Checks.}
\phantomsection\label{sec:hash_checks}
Since updates and compute are replicated across model replicas (i.e., DDP
ranks), we periodically hash model weights across replicas to assert that they
remain bit-identical. These checks primarily catch silent data corruption (SDC)
\citep{dixit2021sdc} from defective GPUs --- specifically, errors originating in arithmetic logic and pipeline registers, which unlike DRAM and SRAM are not covered by ECC
protection. E.g., during one run we identified a single
machine producing silently corrupted arithmetic,
surfacing as anomalous global gradient norms peaking at $\sim 10^{6}$. The
same machine had participated in earlier runs but caused only subtle
symptoms there, owing to lower activation magnitudes. After isolation,
training continued through a short correction period.
\smallskip
Beyond their original purpose, the hash checks have also been valuable for
surfacing subtle logical bugs in our distributed training implementation (data
races, collective communication bugs, replica divergence) that would otherwise
have manifested as slow, hard-to-debug drift. For instance, they caught several
sources of non-determinism in Muon's Newton--Schulz iterations --- whose extra
norm reductions and matmuls offer more opportunities for non-determinism than
AdamW --- which we resolved by pinning the relevant kernels to deterministic
algorithms. They also surfaced a subtle drift caused by our checkpointer
silently inserting an extra cast in the optimizer state-dict path, leaving DDP
rank 0 with \texttt{FP64} beta factors, while all other ranks used \texttt{FP32}.
\end{notebox}

\paragraph{Data Loading and Checkpointing.} We stream training data rather than reading it from local disk. Streaming is handled by Blender, our internal data service, which mixes multiple sources according to specified mixture weights and curriculum, and exposes a gRPC API for fetching batches with consistent global composition and synchronization across workers. On the training side, a sidecar process prefetches batches from Blender and hands them to the main process through shared memory; the main process then copies them to the GPU via pinned memory. Checkpoints are
persisted primarily to S3, with writes distributed across all nodes
within a single model replica to make use of the available write
bandwidth. Reliable S3 interaction at this scale required substantial
engineering effort to handle request throttling and transient errors
gracefully on both the read and write paths. On reads, to avoid
throttling at restart, only a single replica fetches the checkpoint from
S3 and broadcasts the weights to the remaining ranks via RDMA.

\subsubsection{Training Stability}

Pre-training \lagunam surfaced several stability issues that required
iteration on our recipe. The lessons from those investigations were carried
into \lagunaxs from the start, and \lagunaxs pre-training proceeded without
encountering further stability issues.

\paragraph{Expert Collapse.} To align the effective weight decay between matrix parameters (updated by Muon) and non-matrix parameters (updated by AdamW~\citep{loshchilov2019adamw}), we adopt Moonlight-style learning-rate scaling~\citep{liu2025muon}. With this, Muon runs at AdamW-scale LRs, removing the order-of-magnitude mismatch in effective WD. In preliminary experiments, we found that Muon without this rescaling and with standard weight decay coefficients caused expert collapse beginning at around 450B tokens into training and propagating layer-by-layer through the first three MoE layers until the training diverged.

\paragraph{Precision of the LM Head Input-gradient All-reduce.}\phantomsection\label{sec:training_stability_grad}

A critical observation for the training stability of \lagunam was the precision of the input-gradient all-reduce for the LM head. By default in our training codebase, the LM head runs under mixed precision
\citep{micikevicius2018mixed}, with \texttt{BF16} gradients on its inputs; under
column-wise tensor parallelism those input gradients are also all-reduced
across ranks in \texttt{BF16}, simply inheriting the gradient dtype. We have no optimization pressure against pre-softmax logit drift --- no z-loss~\citep{zoph2022stmoedesigningstabletransferable, chowdhery2023palm}, and RMSNorm does not subtract the mean --- so the logits can freely grow during training. In \lagunam this manifested as a strongly positive logit drift. With growing activations the input-gradient all-reduce becomes the dominant source of numerical error and propagates to the rest of the model, destabilizing training. We therefore enforce the LM head
input-gradient all-reduce in \texttt{FP32} while keeping the layer
tensor-parallel, resolving this numerical instability.

\paragraph{Padding in MoE Routing.} A separate observation during
\lagunam training was that, despite sequence packing, $\sim$5\% of training tokens were padding, and these were both routed and included in the
load-balancing loss. Because we mask the language-modeling loss on padding
tokens, the padding embedding is non-learnable; combined with the fact that
padding tokens flow through attention without mixing with surrounding
tokens, every padding token arrives at the router with the same
representation. As a consequence, all padding tokens in a batch are routed
to the same expert at the same time, potentially saturating that expert. For \lagunaxs\!\!, we added an
option to skip routing and load-balancing of padding tokens; further
ablations confirmed this is preferable from a routing-stability perspective.

\subsection{Pre-training Data}
\label{sec:data}

We trained both \lagunam and \lagunaxs from scratch on a pre-training corpus sampled from a pool of $\sim$27T unique tokens. Our training data spans a broad collection of sources, including large-scale web corpora, code repositories, curated educational datasets, academic text, and synthetically generated data. In total, both models were trained on more than 30T tokens.

Optimal dataset design for such long training regimes differs from those for shorter training horizons. Recent work on mixture pre-training under data constraints shows that the optimal mixture depends jointly on target data size, mixture ratio, model size, and compute budget~\citep{sedova2026mixtureconstraints}. Hence typical data strategies that optimize for high precision by aggressively removing noisy documents to improve average quality (e.g.\ in~\citep{fineweb,dclm}) are not optimal for larger token budgets. 

We made this observation when training \lagunam\!\!, where we used a high-precision pipeline with a manually designed mixture, which exposed two bottlenecks: (1) excessive repetition in high-value subsets and (2) suboptimal allocation of data budget across different sources. We confirmed these observations also afterwards in targeted smaller scale experiments. The challenge shifted from maximizing precision under scarcity to controlling repetition and diversity under long-horizon training. 

For \lagunaxs we addressed these challenges through three main efforts:
\begin{itemize}
  \item \textbf{High-recall web data.} We shifted from a predominantly high-precision web pipeline toward a substantially higher-recall pipeline designed to preserve diversity while maintaining quality.
  \item \textbf{Synthetic data at scale.} We expanded the amount of usable training signal through targeted synthetic rephrasing at scale.
  \item \textbf{AutoMixer.} We replaced static manually designed mixtures with a custom automated data mixture process.
\end{itemize} 

\subsubsection{Web Data}\label{sec:web_data}

Our web data pipeline for both \laguna models ingests large scale web data as raw HTML and we use a custom parser that maximizes data recall and optimizes boilerplate removal and the parsing of technical content. Further data cleaning is performed before language identification. This pipeline leverages GlotLID~\citep{kargaran-etal-2023-glotlid} with pycld2 as a fallback to filter English content. Snapshot-level fuzzy deduplication is preferred because it gives higher performance on knowledge benchmarks. Indeed, we observed that similar web pages across snapshots are statistically more likely to contain relevant facts.

Large-scale web data forms the broadest component of the pre-training corpus, providing coverage across topics, domains, languages, formats, and writing styles that would be infeasible to curate or enumerate manually. For large token budgets our objective is to maximize recall over useful documents, particularly within the high-quality segments, while intentionally retaining enough mid- and lower-quality material to preserve long-tail diversity and sustain scaling performance.

\begin{notebox}[Model Factory Component: Spark for Data Pipelines]
We consolidate every data processing step to run on Spark, and the intermediate results of each data processing step are registered as a Dagster asset. At steady state the system processes on the order of $2{\times}10^{13}$ tokens per day.
\end{notebox}

\paragraph{Data Labelling.}
One of the core challenges in large-scale web curation is defining a sufficiently stable and general criterion for judging document quality.
We frame the core question as how to define \emph{useful} versus \emph{non-useful} content in a way that remains consistent across domains, formats, and scales of training.

To construct labels for filtering, we model document quality along two complementary axes: a \emph{noise} axis ($N$) that captures whether a document is primarily noise or low-information content, while an information axis ($I$) captures whether the document contains educational, informational, or broader pre-training value. We use an integer range in $[0, 5]$ for both axes. This separation matters because many web documents are imperfect but still useful; a page may contain excessive boilerplate or formatting artifacts while still retaining substantial informational or educational value.

Each annotated document is additionally mapped onto the contribution scale defined in \Cref{tab:web_ground_truth_anchors}.
Particular attention is given to the ambiguous low-quality region, especially documents in the $[0, 2]$ range of the contribution scale, where weak, noisy, and partially useful content overlap.
We densely annotate this region within the $N \times I$ space to better separate truly unusable documents from imperfect but recoverable data.
This calibration provides a substantially more stable target for downstream model-based quality scoring: aggressively filtering clear noise while preserving lower-confidence documents that still contribute useful long-tail training signal.

\begin{table}[h]
  \centering
  \scriptsize
  \setlength{\tabcolsep}{4pt}
  \renewcommand{\arraystretch}{1.25}
  \begin{tabular}{@{}cp{0.25\linewidth}p{0.67\linewidth}@{}}
    \toprule
    \textbf{Score} & \textbf{Anchor} & \textbf{Pre-training interpretation} \\
    \midrule
    0 & Useless & Almost no training value; contributes no meaningful capability signal. \\
    1 & Clearly harmful & Overall content is harmful to training, e.g.\ severe logic errors, toxic patterns, or misinformation. \\
    2 & Borderline-negative & Uncertain but leaning negative; quality is too low or inconsistent to support effective pre-training. \\
    3 & Borderline-positive & Uncertain but leaning positive; core value is graspable and benefits outweigh weaknesses. \\
    4 & Confirmed beneficial & Clear positive impact despite some noise, such as useful logic, knowledge, or language patterns. \\
    5 & Flawless contribution & Minimal noise and exceptional expected value for improving model capabilities. \\
    \bottomrule
  \end{tabular}
  \vspace{5pt}
  \caption{Universal anchors used to build the web ground truth. The
  $N \times I$ grid is used to make the low-score boundary precise rather
  than treating all imperfect web pages as discardable.}
  \label{tab:web_ground_truth_anchors}
\end{table}

\paragraph{Ranking rather than Filtering.}
Rule-based filters are useful for removing obvious artifacts, but exposed a  precision/coverage trade-off we observed in training \lagunam\!\!: broader rules removed more noise at the cost of discarding useful mid-quality documents, while highly precise rules covered only a small fraction of the noisy tail.
For \lagunaxs\!\!, the primary rejection path is therefore model-based and deliberately conservative. The pipeline removes documents only when we're confident they are pure noise. For the remaining documents, quality is treated as a ranking signal rather than fully filtered out.
Documents are then sorted by a continuous contribution score, allowing us to fill data allocation quota by this ranking.

While ranking according to one general quality dimension would be ideal, we found that a single dimension for classification is too coarse and not accurately predictable enough. Hence we decompose document quality into a set of independently learnable properties, which we then recombine into a composite contribution score.

The ranking pipeline therefore leverages model-based annotations to remove high-confidence noise, assigns continuous quality signals, and samples from corresponding quality buckets. The document tags are produced with Propella~\citep{idahl2026propella}, an open multi-property
document annotation model that exposes separate dimensions rather than a single quality scalar. We keep Propella's native rating ranges, use a
PCA-informed subset of its dimensions to de-correlate
signals, and then form the composite score that is used both for filtering and sampling.
Through empirical filtering and training experiments, we found that pre-training usefulness is best modeled through multiple complementary dimensions, including content quality, educational value, information density, content integrity, content ratio, and commercial bias.

Our composite score completely filters out $25.8\%$ of web samples, while recovering roughly $34\%$ of high-quality documents previously excluded by our previous static rules and lexical-based classifiers. Note that surviving the filtering does not mean the sample is necessarily taken into training. Instead they just become eligible to being sampled according to their contribution score.

\begin{figure}[h]
  \centering
  \includegraphics[width=\textwidth]{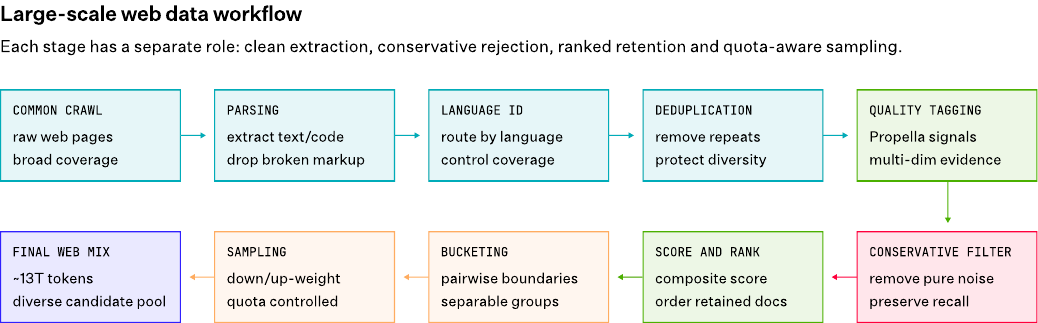}
  \vspace{5pt}
  \caption{High-level workflow for large-scale web data. Early stages extract,
  route, and deduplicate Common Crawl documents. Quality tagging supplies
  multi-dimensional evidence, filtering conservatively removes pure noise,
  composite scoring ranks retained documents, and bucketing plus sampling turns
  the ranked pool into a quota-controlled web mixture.}
  \label{fig:web_pipeline_workflow}
\end{figure}

\begin{figure}[h]
  \centering
  \includegraphics[width=\textwidth]{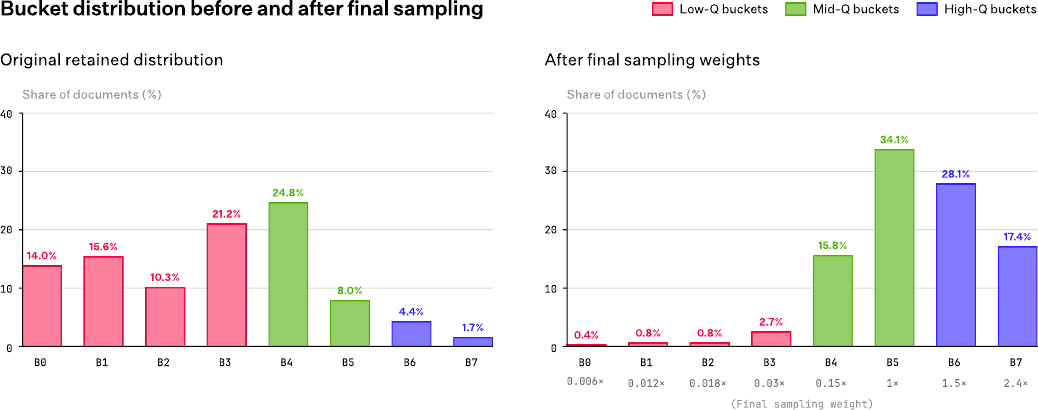}
  \vspace{5pt}
  \caption{Document-level bucket distribution before and after applying the
  final sampling weights. Lower-quality buckets are progressively downsampled, while most sampling weight is allocated to higher-quality regions [B4-B7]. This preserves diversity while still prioritizing higher-value documents during training.}
  \label{fig:web_sampling_distribution}
\end{figure}

\paragraph{Bucket Calibration.}
After filtering, retained documents are grouped by the composite score for
sampling. For sampling, we define bucket boundaries by
blind pairwise comparisons around candidate thresholds, so documents across a
boundary are distinguishable by expected pre-training value. This produces
variable-sized but more separable buckets. Finally, \Cref{fig:web_sampling_distribution} shows the final sampling
distribution across quality buckets.

\subsubsection{Synthetic Data}\label{sec:synth_data}

\providecommand{\Pset}{\mathcal{P}}
\providecommand{\Sset}{\mathcal{S}}
\providecommand{\Mset}{\mathcal{M}}
\providecommand{\Oset}{\mathcal{O}}

We use synthetic data to complement our organic data sources. Synthetic data builds on top of those sources rather than replacing them, regularizing presentation, filling under-represented formats, and exposing structure (plans, rationales, question-answer surfaces, etc.). In \lagunaxs it carries $\sim\!13\%$ of the mix across all stages, drawn from a pool of $\sim\!4.4$T generated tokens, spanning seed-heavy form rephrasing~\citep{maini2024wrap, su2025nemotroncc} and expensive pipeline-heavy compositional distillation~\citep{abdin2024phi4, gunasekar2023textbooks}. We apply these pipelines across high-signal STEM and code domains and extend them into the broader data mix, so synthetic data enters early and runs consistently across all training stages.

We take a modular approach to synthesis: each pipeline composes from a small set of reusable components among six types:
\begin{enumerate}
  \item A set of inputs $\Sset$.
  \item Metadata attached to each transformation in the pipeline $\Mset$.
  \item A generator (LLM or a chain of LLMs) $G$ we sample from to produce outputs $o \sim G(s, m)$ for some sample $s$ and metadata $m$.
  \item Filtering functions $f$.
  \item Validation function $V$.
  \item Pre- and post-processing functions $\mathrm{pre}$ and $\mathrm{post}$.
\end{enumerate}

This allows us to express each pipeline $P$ as a composition of components, like below. 
\begin{equation*}
P \;=\; \mathrm{post}\,\circ\,f_{n}\circ G_{n}\circ\cdots\circ f_{1}\circ G_{1}\,\circ\,\mathrm{pre}.
\end{equation*}
In practice we draw a sample with metadata $(s, m)$ from the input pool, draw a new sample conditioned on it $o \sim G(\cdot \mid s, m)$, and keep $o$ if $f$ accepts. The resulting synthetic corpus is the set of outputs generated by repeatedly sampling inputs from $\Sset$.

\begin{figure}[h]
\includegraphics[width=\textwidth]{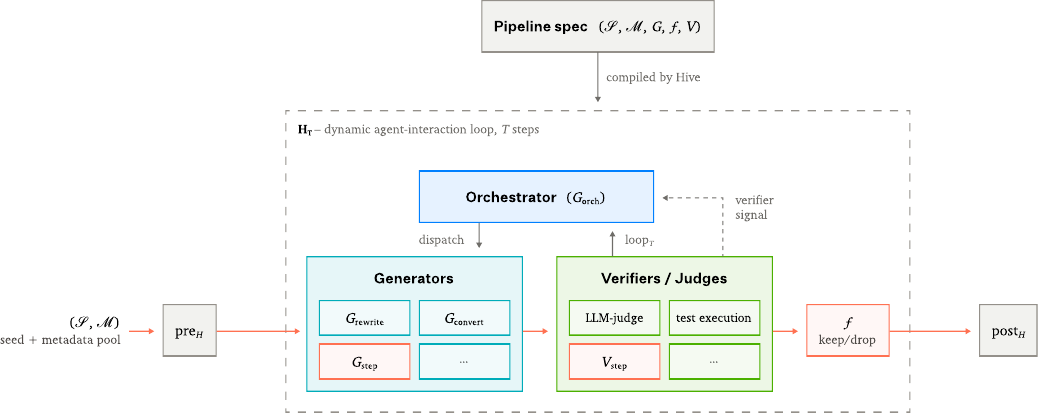}
\caption{\textbf{Hive runtime.} A declarative pipeline spec $(\mathcal{S},\mathcal{M},G,f,V)$ is compiled into a single
$H_T$ loop. A seed and metadata pool $(\mathcal{S},\mathcal{M})$ feeds $\mathrm{pre}_H$; $\mathrm{pre}_H$ and
$\mathrm{post}_H$ wrap the loop, while inside it an orchestrator dispatches across a library of generators and verifiers
over $T$ rounds and $f$ gates keep or drop outputs.}
\label{fig:hive}
\end{figure}

\begin{notebox}[Model Factory Component: Hive for Synthetic Data]
To enable defining and running such pipelines easily, we built \emph{Hive} as a component of the Model Factory (\Cref{sec:factory}), a configurable framework that runs every pipeline as
\begin{equation*}
P \;=\; \mathrm{post}_{H}\,\circ\, H_{T}\,\circ\,\mathrm{pre}_{H},
\end{equation*}
where $H_T$ is a dynamic agent-interaction loop over orchestrators, generators, judges~\citep{wu2023autogen, zheng2023judging}, and per-step early-exit gates given a common workflow. See \Cref{fig:hive} for an illustration of the runtime. Iterating on a synthetic data generation pipeline is a configuration change rather than a re-implementation.
\end{notebox}

Generating useful and diverse synthetic data is a challenging task. For our pipelines, we follow two broad principles:
\begin{itemize}
  \item \textbf{Match pipeline complexity to teacher capability.} A task the teacher cannot solve in one shot becomes a source of biased generation, so we either decompose or strengthen the teacher.
  \item \textbf{Help LLMs through metadata.} We hand our generator chain $G$ anything we already know about the output or how to achieve it through supplying metadata, reducing the challenge on the generator alone.
\end{itemize}

Following these principles, we define a series of synthetic data pipeline strategies that we used for \lagunaxs\!\!. Each strategy balances between sometimes competing axes of \emph{Grounding} (staying faithful to the information content in the seed/input data), and \emph{Entropy} (diversity, injecting randomness and variation into the input data). Through composition of stages, filters, generators, and verifiers, we can define elaborate generation pipelines optimizing for both~\citep{yu2023diversity, lee2023diversity}. We describe some of these strategies in \Cref{tab:synth_shapes}.

\begin{table}[h]
\centering
\footnotesize
\renewcommand{\arraystretch}{1.25}
\setlength{\tabcolsep}{4pt}
\begin{tabular}{@{}p{0.14\linewidth}p{0.40\linewidth}p{0.30\linewidth}p{0.08\linewidth}@{}}
\toprule
\textbf{Shape} & \textbf{Composition} & \textbf{What it does \,/\, examples} & \textbf{Tokens} \\
\midrule
Form-rewrite &
$P=\mathrm{post}\circ f_{\text{quality}}\circ G_{\text{rewrite}}\circ\mathrm{pre}$ &
Single-call pass conditioned on $\Mset$ (mode, voice, structure); content inherited from $\Sset$. Used for: multi-mode rephrasing of web and \textsc{STEM} docs; code rephrased into code with natural-language. &
$\sim$$10^{12}$ \\
Cross-domain transducer &
$P=\mathrm{post}\circ f_{\text{format}}\circ G_{\text{convert}}\circ\mathrm{pre}_{\text{seed}}$ &
Casts a seed across modalities; $\mathrm{pre}$ enforces seed suitability. Used for: math\,$\leftrightarrow$\,code, code-language porting. &
$\sim$$10^{10}$ \\
Multi-stage cascade &
$P=\mathrm{post}\circ f_{\text{keep}}\circ G_n\circ\cdots\circ(f[V\!\geq\!\tau]\circ V\circ G_k)\circ\cdots\circ G_1\circ\mathrm{pre}$ &
Each $G_i$ transforms the previous output or emits metadata for later stages; the parenthesized $V$-gate prunes intermediate outputs. Used for: academic and technical content adapted across multiple educational formats, textbook synthesis, diff-conditioned coding task generation, grounded QA generation. &
$\sim$$10^{11}$ \\
Multi-turn rollout &
$P=f_{\text{bounds}}\circ \mathrm{loop}_{T}
\left(\bigcirc_{i=1}^{k}(G_{\text{step}}^{(i)}\circ G_{\text{orch}}^{(i)})\right)
\circ G_{\text{init}}\circ f_{\text{relevance}}$ &
Closed-loop interaction; orchestrator decides flow and termination. Used for: stacktrace- and raw-code-grounded multi-turn chats, multi-turn math resolution, iterative eval-based doc evolution. &
$\sim$$10^{10}$ \\
\bottomrule
\end{tabular}
\vspace{5pt}
\caption{Pre-training synthetic data pipeline strategies. Token counts shown as order of magnitude of contribution to our pre-training corpus.
}
\label{tab:synth_shapes}
\end{table}

\subsubsection{AutoMixer}
\label{sec:automixing}
Pre-training data composition has a substantial impact on downstream model capabilities, with prior work showing that domain mixture proportions can significantly affect model quality~\citep{doremi}. Optimizing data mixtures manually becomes intractable as the number of datasets, capability trade-offs, and training constraints grows. 
Traditional approaches based on heuristics, qualitative strategies, or manual ablations are expensive, low-dimensional, and slow to iterate on, particularly at modern pre-training scales~\citep{data_mixing_laws}.
Recent work has therefore explored learned, proxy-based, or benchmark-driven approaches for pre-training data optimization and mixture design~\citep{datacomp_lm,olmix,mde,regmix}.

Our framework is inspired by these directions, but adapted to our setting. To solve the dataset mixing problem, we developed \emph{AutoMixer}, a framework for automated data mixture optimization. AutoMixer operates by training a swarm of proxy models, each trained on a different data composition.
For each data ablation run, we have trained $\sim$ 60 proxy models, each as a $\sim$0.5B parameter MoE model on $\sim$ 60B tokens sampled from different mixtures across our pre-training corpus.

\begin{figure}[h]
\centering
\includegraphics[width=0.7\linewidth]{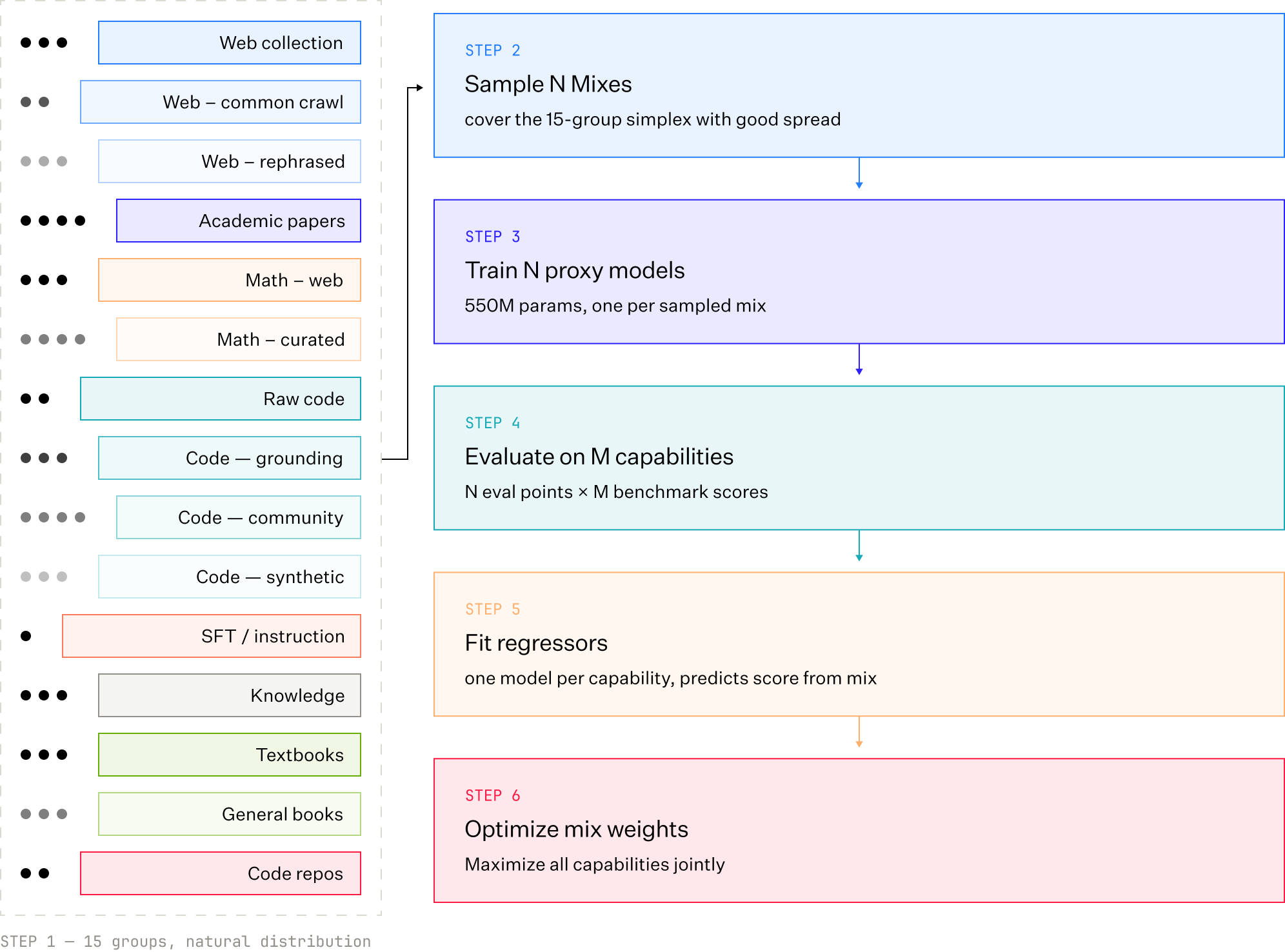}
\vspace{5pt}
\caption{AutoMixer pipeline. A swarm of proxy models is trained under controlled mixture perturbations. Capability evaluations are used to fit surrogate regressors, which are subsequently optimized under regularization and practical mixture constraints.}
\label{fig:automixer_framework}
\end{figure}

The explored corpus spans more than 50 heterogeneous dataset groups, including general web corpora, curated educational text, academic papers, raw code, grounded code data, synthetic code data, mathematical web content, and conversational and knowledge-focused datasets. The overall framework is illustrated in \Cref{fig:automixer_framework}.

The core idea behind AutoMixer is to learn a surrogate mapping:
\[
\mathcal{M}: x \rightarrow y
\]
where $x \in \Delta^d$ denotes a data mixture over $d$ dataset groups, and $y \in \mathbb{R}^k$ denotes downstream evaluation metrics across $k$ capability groups. Through learning $\mathcal{M}$, AutoMixer learns how changes in mixture proportions affect downstream capabilities and we can then directly optimize over the learned surrogate.

We start the exploration of the data mixture search space by sampling over a manually designed prior mixture $x_0$. Candidate mixtures are sampled as:
\[
x \sim \mathrm{Dirichlet}(\alpha x_0)
\]
subject to additional constraints:
\[
\|x - x_0\|_1 < \epsilon
\]
This allows efficient exploration of meaningful regions of the mixture space while avoiding unrealistic or degenerate
configurations. For each capability group, we train a surrogate regressor:
\[
f_j(x) \approx y_j
\]
where $x$ is the mixture vector and $y_j$ is the downstream evaluation metric for capability group $j$.

Capabilities are grouped into coding, mathematical reasoning, STEM knowledge, commonsense reasoning, and general knowledge. We use a small set of pre-training benchmarks as a proxy for downstream capabilities.

The independent
regressors allow the framework to recover capability-specific sensitivities and trade-offs. A simplified linearized form can be written as:
\[
\hat{y}_j = \beta_j^\top x + b_j
\]
although in practice the final implementation uses non-linear surrogate models.

The learned signals recover intuitive relationships between subsets and downstream capabilities. For instance, we observed synthetic and curated
code data to strongly improve coding evaluations, while conversational and knowledge-centric corpora improve commonsense reasoning. Importantly, the framework recovers not only expected high-level relationships, but also substantially finer-grained interactions between subsets.

Once the surrogate regressors are trained, mixture optimization is formulated as:
\[
\max_x \sum_{j=1}^{k} w_j f_j(x)
\]
subject to:
\[
\sum_i x_i = 1
\]
\[
x_i \ge 0
\]
\[
\|x - x_0\|_1 < \epsilon
\]
To avoid unrealistic shifts toward a small number of dominant sources, we regularize the optimization toward the
baseline prior:
\[
\mathcal{L}(x)
=
-\sum_j w_j f_j(x)
+
\lambda D_{KL}(x \| x_0)
\]
where $x_0$ is the baseline mixture, $D_{KL}$ is the KL-divergence regularization term, and $\lambda$ controls deviation from the prior. In smaller scale ablations, we validated that the optimized mixture improves strongly over the manually designed prior mixture. \Cref{tab:automixer_results} summarizes the
downstream performance changes across both optimization benchmarks focused on coding, and held-out evaluations during a small-scale experiment on a 3B parameter model for 1.5T tokens.

We see that gains generalize to held-out benchmarks that were not directly optimized against, suggesting that the learned surrogate optimization generalizes meaningfully. We also observe modest regressions on several commonsense-oriented benchmarks, which is not surprising given the optimization objectives we chose.

\begin{table}[t] \centering \small \begin{tabular}{lll} \toprule Category & Benchmark & Relative Change \\ \midrule \multirow{7}{*}{Optimization Targets} & HumanEval+ & +43\% \\ & MBPP+ & +15\% \\ & Crux-I & +54\% \\ & Crux-O & +48\% \\ & MultiPL-E & +27\% \\ & GSM8K & +41\% \\ & MMLU & +5\% \\ \midrule \multirow{4}{*}{Held-out Benchmarks} & MATH & +25\% \\ & LiveCodeBench & +39\% \\ & APTBench-4k & +35\% \\ & BigCodeBench & +16\% \\ \midrule \multirow{4}{*}{Commonsense Trade-offs} & ARC-C & -6.8\% \\ & WinoGrande & -1.4\% \\ & PIQA & -0.9\% \\ & HellaSwag & -1.3\% \\ \bottomrule \end{tabular}
\vspace{5pt}
\caption{ Performance impact of the optimized AutoMixer data mixture. Large gains are observed on optimization targets, particularly coding and mathematical reasoning benchmarks. Improvements also generalize to held-out evaluations not used during optimization, while modest regressions are observed on several commonsense-oriented tasks. } \label{tab:automixer_results} \end{table}

\begin{table}[h]
\centering
\small
\begin{tabular}{lc}
\toprule
Data group & Mix weight \\
\midrule
Raw code & 30.6\% \\
Web & 25.2\% \\
Synthetic/code-text & 25.4\% \\
Math & 9.0\% \\
Knowledge & 6.6\% \\
Instruction-like & 1.4\% \\
Academic papers & 1.1\% \\
Books & 0.7\% \\
\bottomrule
\end{tabular}
\vspace{5pt}
\caption{Aggregated pre-training data mix used for \lagunaxsshort\!\!.}
\label{tab:laguna_xs_mix}
\end{table}

AutoMixer plays a key part in our data strategy for pre-training. The final optimized mixture used for \lagunaxsshort\ is shown in \Cref{tab:laguna_xs_mix}.
Relative to the data mixture used for \lagunam\!\!, the allocation defined by AutoMixer used by \lagunaxs shifts substantially toward broader web coverage, synthetic/code-text data, and mathematically oriented corpora, while still preserving a strong code-heavy foundation.

\section{Post-training}\label{sec:post-training}

Post-training for \lagunaxs and \lagunam picks up after pre-training, including context-length extension (\Cref{sec:training_recipe}), and is divided into three stages:

\begin{itemize}
    \item \textbf{Mid-training:} The largest stage by unique token count, containing roughly $60$B tokens in a broad mix of instruction data spanning chat, reasoning, and agentic coding.
    \item \textbf{Supervised fine-tuning (SFT):} Three epochs of roughly $40$B tokens each, focused primarily on agentic coding.
    \item \textbf{Reinforcement learning (RL):} Using only verifiable rewards and performed online with CISPO.
\end{itemize}

We iterate on data, hyperparameters and some architectural choices on top of \lagunam\!\!, which was available earlier. Due to the fast turnaround time in adapting pre-training changes from \lagunamshort to \lagunaxsshort\!\!, the smaller model was available shortly after, and we proceeded with parallel post-training of both models. Apart from hyperparameter changes and minor dataset fixes, the post-training recipe stayed the same between \lagunam and \lagunaxsshort.

\begin{notebox}[Model Factory: Research is Self-Service]
The initial imitation learning stages for \lagunaxs were run without any need for human synchronization and performed by a member outside of the core post-training team. This was made possible because the underlying tooling is shared across the whole company, and all training runs are tracked with their full lineage. Training runs can be easily replicated and adapted by any researcher in a self-service manner.
\end{notebox}

\subsection{Special Tokens and Templates}

\subsubsection{Special Tokens}

We introduce new XML-like special tokens to be used to format multi-turn chats in the mid-training stage: \texttt{<assistant>}, \texttt{</assistant>}, \texttt{<think>}, \texttt{</think>}, \texttt{<tool\_call>}, \texttt{</tool\_call>}.
Their embeddings are randomly initialized before pre-training and remain unaltered until the mid-training stage.
For the smaller \lagunaxs model, random initialization resulted in successful
mid-training; however, in \lagunam the mismatch between special and regular token embeddings caused gradient spikes, dead experts, and broader training
instabilities.

We address this with initialization and a short warmup phase before mid-training.
For initialization, we use subtoken averaging~\citep{sachidananda2021efficient, dobler2025tokendistillation} --- each new token's embedding is initialized as the mean of its constituent subtokens' embeddings.
For example, \texttt{<think>} is initialized as the mean of \texttt{<th}, \texttt{ink}, \texttt{>}.
This gives the token a semantically grounded starting point consistent with the surrounding embedding space. We then run a 100-step warmup phase where the full network except input embeddings and LM head layers is frozen, allowing the new token representations to adapt before the main post-training phases begin.
We found this strategy produces the most stable training dynamics and the lowest tool call formatting error rates from the start of training.

\subsubsection{Formatting and Structured Outputs}\label{sec:reasoning-toggle}
We adopted chat and structured-output formats that align with open-source
conventions: reasoning blocks use \texttt{<think>}/\texttt{</think>} tags, and
tool calls follow an XML-like format compatible with the GLM~\citep{glm5team2026glm5vibecodingagentic} series.

To control reasoning, the chat template exposes an \texttt{enable\_thinking} flag.
When enabled, the chat template prefixes an opening \texttt{<think>} tag just before generation, allowing the model to either close the \texttt{<think>} block immediately or generate additional thinking tokens.
During the SFT phase, however, we consistently omit empty reasoning blocks and the model learns to emit reasoning tokens whenever the flag is enabled.
Conversely, when this flag is disabled, the chat template emits only a closing \texttt{</think>} tag.
We train with \emph{persistent thinking history}, where all prior reasoning blocks remain visible in the context during each generation step.

\smallskip

When integrating with open-source tool parsers, we discovered a number of failure cases due to the fact that these parsers are tailored to the set of features specific to their corresponding chat templates.
Several edge cases, not present in other standard usage, made those failure cases often invisible or at least under-reported.
For example, the parsers were sensitive to multi-token streaming deltas: when a single delta contained more than one token, the parser could emit invalid tool calls.

We upstreamed improved reasoning and tool parsers to vLLM,\footnote{\url{https://github.com/vllm-project/vllm/pull/35208}} derived from its
\texttt{deepseek\_v3} reasoning parser and \texttt{glm45} tool parser, to fix two issues:

\begin{itemize}
    \item \textbf{Reasoning mode detection.} In streaming mode, the
    \texttt{deepseek\_v3} reasoning parser decides whether a given turn's
    reasoning has ended by scanning the entire prompt for a closing
    \texttt{</think>}. Our parser instead scans only the current turn's content,
    so prior reasoning spans in the conversation history do not short-circuit
    the check.
    \item \textbf{Deltas spanning block boundaries.} The original parsers
    assumed each streamed delta contained exactly one token. This simplified
    logic because the parser did not need to handle transitions across block
    boundaries (e.g., reasoning $\to$ content $\to$ tool calls). In practice,
    deltas can contain multiple tokens --- for example, under speculative decoding
    or when vLLM's single-slot output collector merges accumulated deltas
    because the producer outruns the consumer. Before our fix, any content
    trailing a boundary token was silently dropped, producing behavior identical
    to invalid special-token usage.
\end{itemize}

\subsubsection{TITO and Chat Template Alignment}

We use a token-in, token-out (TITO) API design for the actors used in our RL training.
This ensures that token IDs are stable across multi-turn interactions, avoiding mismatches in the token ID representation of model outputs.
While this approach resolves significant issues in multi-turn, agentic RL training, it introduces the possibility of misalignment between the multi-turn trajectories produced by such a system and the chat template used for production deployments. Subtle differences --- such as a missing newline or a trailing space --- can lead to significant degradation in model performance when the RL-trained model is deployed.

To prevent divergence while still allowing independent iteration of the RL renderer and the production chat template, we introduce a \texttt{render\_assistant\_messages\_raw} flag. When enabled, the chat template inserts the raw rollout tokens verbatim into the assistant message blocks. After every generation step we render the conversation history with this flag enabled and assert that the resulting string exactly matches the decoded prefix stored during rollout. This invariant guarantees that the production template and its logic identically match the behavior seen during RL, eliminating the risk of deployment-time mismatch.

\subsection{Mid-training}

The mid-training stage uses an effective batch size of $128$, a maximum sequence length of $131072$ tokens, and a cosine learning-rate schedule with a 50-step warmup, a peak learning rate of $1{\times}10^{-5}$, and a final learning rate of $2{\times}10^{-7}$. We use Muon throughout. We pack sequences shorter than the maximum sequence length into a single microbatch. We run for a single epoch.

The data corpus used in this stage covers the following categories:

\begin{itemize}
    \item \textbf{General chat}: open-domain question answering, writing, multi-turn dialogue, instruction-following, and long-context interactions;
    \item \textbf{Reasoning}: mathematical, scientific, and logical reasoning trajectories, with explicit chain-of-thought wrapped inside special tokens;
    \item \textbf{Coding and agent}: repository-level code, synthetic code, tool-calling, and long-horizon agentic trajectories targeting use cases for code repositories and terminal environments.
\end{itemize}

We find it critical to tune the following properties of our mid-training data mix:

\begin{itemize}
    \item Presence of tool calls, count and variety of the tools provided in each trajectory, and average number of tool calls per trajectory.
    \item Presence of reasoning, and the ratio of samples with reasoning vs. samples without reasoning.
    \item Length of reasoning and difficulty of question, measured by the difficulty labels of our input task sets.
    \item Total turn count per trajectory, and tokens per assistant and user turn.
\end{itemize}

In total, the mid-training stage consumes approximately $60$B tokens. The mix is deliberately broad: general chat maintains conversational and instruction-following behavior, reasoning data teaches explicit thinking traces and difficult problem solving, and coding-and-agent data teaches repository-level tool use, terminal workflows, and long-horizon software engineering behavior.

In early iterations of our mid-training mix, we observed trajectory-length and tool-availability-dependent failure modes in tool calling and reasoning.
For example, on some domain-specific evaluations, the model would either omit reasoning or generate degenerate reasoning spans when tools were not provided.
Balancing these features across domains resolved the issue.

In terms of data mixture, during mid-training we spend approximately 40\% of the training budget on logic and reasoning, 30\% on coding-and-agent data, and 30\% on general chat. In contrast, during SFT the mix is heavily weighted toward agentic coding: approximately 85\% of tokens are agentic trajectories, with the remaining 15\% shared across the other domains.

\subsection{Supervised Fine-tuning}

The SFT stage shares the same batch size, sequence length, packing, learning-rate schedule and optimizer as mid-training. We run for three epochs of $40$B tokens each, with early stopping based on evaluation scores. 

The dataset for this stage consists of four main parts:
\begin{enumerate}
    \item \textbf{Agentic coding without reasoning.} Single-turn trajectories generated by an open-source teacher model. This component accounts for approximately 30\% of the training mixture by tokens.
    \item \textbf{Agentic coding with reasoning.} A corpus with the same number of samples and a similar source distribution as the previous component, but augmented with reasoning traces. Due to the reasoning, this component constitutes approximately 45\% of the mixture by tokens.
    \item \textbf{Math corpus.} A relatively small agentic mathematics corpus (\(\sim\)3\% of the mixture by tokens), containing tasks both with and without reasoning traces. The data was collected from several open-source datasets and filtered for verifiability, retaining only problems with numeric (rather than symbolic) answers. We additionally removed overly easy samples based on solve rates from open-weight LLMs.
    \item \textbf{Non-agentic samples.} Approximately 22\% of the mixture by tokens. This component is included to mitigate forgetting of non-agentic capabilities.
\end{enumerate}

The agentic coding samples are collected from multiple sources, including open-source datasets and publicly available code repositories. A large portion of our SFT dataset is generated synthetically in-house, sharing the same pipeline that generated synthetic code environments.

In general, the tasks in the agentic coding corpus focus specifically on software engineering capabilities to operate with code repositories or directly on the text-based terminal.

\subsubsection{Synthetic Code Environments}

A large portion of our agentic software engineering training tasks is produced by an internal pipeline that turns real-world \texttt{git} commits from public repositories into verifiable training tasks.
Each commit becomes a task whose problem statement, repository checkout, and hidden test patch are taken directly from the commit, with the commit diff kept as a gold solution. Tasks are filtered with a two-sided correctness check that requires the gold solution to pass the tests and an empty solution to fail them; this removes trivial tests and commits whose tests do not actually exercise the change. We additionally filter on repository popularity and code-quality percentiles, optionally keeping a single task per repository to enforce codebase diversity, retaining on the order of $30$--$60$k tasks from a raw pool of ${\sim}236$k commits. The resulting tasks feed both the SFT mix --- via teacher-generated trajectories, optionally augmented with multiple synthetic system messages for prompt diversity --- and the RL task pool, where the per-repository test suite serves directly as the binary verifier.

\subsubsection{Instruction Following}

A key challenge in agentic SFT is \emph{instruction following} (IF): ensuring the model respects behavioral constraints specified in the system prompt (e.g., tool restrictions, output format rules, persona requirements). Without explicit IF supervision, RL fine-tuning tends to exhibit catastrophic forgetting of these behaviors, and over half of agentic tasks receive zero reward because the model violates system-message constraints before any coding work is evaluated.

\paragraph{Data Augmentation Pipeline.}
We augment agentic tasks with synthetically generated system messages before trajectory generation.
For each task drawn from our agentic software engineering training set, we apply an
EvolInstruct-style~\citep{xu2023wizardlm} generator to produce a set of behavioral requirements
grounded in the task context.
These requirements are appended to the original system message, so that the
resulting trajectories demonstrate a model that simultaneously solves the coding task and satisfies
explicit constraints.
We generate multiple augmented system messages per task and use this approach to produce the final
training dataset at scale.

We developed a dedicated IF judge that decomposes a system message into a list of independent
behavioral requirements and scores each against the model's trajectory.
Judge quality was validated against human-annotated examples and calibrated via prompt tuning;
inter-model agreement was measured across several frontier models.

\smallskip

To track progress on ablation runs, we created an internal evaluation set obtained with the same pipeline, and applied to SWE-bench Verified.
This allowed us to have complementary views of task completion (pass rate) and instruction adherence (follow rate).

Those ablations varied the training data along several axes: the number of requirements, system-message retention (we dropped the original one), the choice of teacher models, the necessity of generated rubrics (we used simple requirement templates in the end), and filtering strictness.

Those ablations allowed us not only to improve on the IF version of SWE-bench Verified, but also on the pass rate directly.

\subsubsection{Multi-harness Training}
To encourage generalization across diverse agent harnesses, our training data includes trajectories from external frameworks such as OpenHands~\citep{Wang_OpenHands_An_Open}, OpenCode\footnote{https://github.com/anomalyco/opencode}, and Mini-SWE-Agent\footnote{https://github.com/SWE-agent/mini-swe-agent}.
Specifically, we augment the supervised fine-tuning mix with 1.3B tokens of multi-harness agentic trajectories spanning a broad range of software engineering tasks.
We intentionally preserve native harness behaviors during data collection ---
including custom subagent spawning, context compaction, planning scaffolds, and
reminder systems --- so the model sees a broad distribution of interaction
patterns. These frameworks vary substantially in trajectory length, tool use,
subagent orchestration, and harness-specific quirks, all of which improved
generalization in internal evaluations.

\subsection{Agentic RL}

Following supervised post-training, we run a large-scale agentic reinforcement-learning phase in which the policy itself drives the deployed agent harness, producing multi-turn trajectories on real coding repositories, terminal sandboxes, and tool-augmented math problems.

The harness driving RL rollouts is the same one we ship in production: the policy interacts through the same tool API, chat template, and orchestration layer that customers exercise, so any change to the deployed harness is also a change the policy is trained against.
Each rollout starts by spinning up a container drawn from the code-execution platform described in \Cref{sec:factory}, which hosts both the agent's working environment for SWE, shell or tool-integrated reasoning tasks and the per-task verifier that produces the reward; the container is torn down once the verifier returns, with several thousand live across the platform at any moment.

\begin{notebox}[Model Factory Component: Code Execution Environments]\phantomsection\label{sec:code_exec}
We maintain a containerized code-execution platform as a primitive to be used in synthetic data generation, evaluations, and providing execution-based rewards in reinforcement learning.

Containers are produced from OCI builds against GitHub sources, with agent-driven generation increasingly replacing handwritten Dockerfiles for repositories that do not build out of the box. This platform spans roughly one million repositories at the moment.
\end{notebox}

\subsubsection{Algorithm}

We train with a token-level REINFORCE surrogate combining importance-sampling-ratio clipping (CISPO~\citep{chen2025minimaxm1}) with a length-weighted leave-one-out~\citep{ahmadian-etal-2024-back} group-relative advantage estimator. 
Moonlight scaling~\citep{liu2025muon} was deactivated during RL for \lagunam\!\!, but used for \lagunaxs\!\!.
We selected this recipe after ablating against other group-based policy-gradient algorithms (GRPO~\citep{shao2024deepseekmath}, GSPO~\citep{zheng2025gspo}), which showed it offered the best combination of final evaluation quality and training stability.

For each prompt we sample a group of $G$ trajectories $\{\tau_i\}_{i=1}^{G}$.
Letting $r_i$ denote the scalar terminal reward of $\tau_i$ and $w_i$ its number of rewarded (assistant-generated) tokens, the length-weighted leave-one-out baseline and advantage are
\begin{equation}
b_i \;=\; \frac{\sum_{j\neq i} w_j\, r_j}{\sum_{j\neq i} w_j}, \qquad A_i \;=\; r_i - b_i.
\end{equation}

With $\pi_\theta$ the current policy and $\pi_{\theta_{\mathrm{old}}}$ the rollout policy at sampling time, the per-token surrogate loss applied uniformly across the assistant tokens of $\tau_i$ is
\begin{equation}
\mathcal{L}(\theta) \;=\; -\,\mathbb{E}\!\left[\, \mathrm{clip}\!\big(\rho_t,\, 1 - c_{\mathrm{low}},\, 1 + c_{\mathrm{high}}\big)\; A_i \; \log \pi_\theta(y_t \mid x, y_{<t}) \,\right],
\quad \rho_t \;=\; \frac{\pi_\theta(y_t \mid x, y_{<t})}{\pi_{\theta_{\mathrm{old}}}(y_t \mid x, y_{<t})},
\end{equation}
with asymmetric clipping bounds $(c_{\mathrm{low}}, c_{\mathrm{high}}) = (1, 4)$, yielding an effective importance-ratio clip of $[0, 5]$ that engages only on heavily off-policy tokens.

\subsubsection{Reward Design}

Rewards are produced by a deterministic chain of checkers applied to every terminated rollout in the following order, with the first failing check determining the reward.

\textbf{Parsing error} ($-0.1$): the rollout contains a malformed tool call or chat-template violation. Applying this penalty only to the last turn suppresses formatting drift without overwhelming the verifier signal.

\textbf{Minimum-steps penalty} ($-0.1$): the agent exits before issuing at least $n_{\min}$ tool calls. $n_{\min}$ was specifically tuned to suppress degenerate short-circuited trajectories that ``give up'' before verifying the solution.

\textbf{Timeout, max steps reached} ($0.0$): a wall-clock timeout or reaching the maximum number of steps zeros the reward without an additional penalty.

\textbf{Binary task verifier} ($1.0$ / $0.0$): the task-specific verifier produces the only positive reward. For SWE-style tasks this is the repository's own unit-test suite run against the agent's patch; for terminal tasks it is the task's bash-side assertion harness; for tool-integrated math problems it is exact numeric-answer match against the gold solution.

\textbf{Tool-error step penalty} ($-0.05$, per-token): for every assistant step whose tool invocation led to the error in tool execution, we apply a small negative reward to exactly the tokens that constituted that step, rather than to the trajectory as a whole. This per-token shaping focuses credit assignment on the failing step and discourages long chains of malformed tool calls before falling back on the terminal verifier.

Shaping is therefore restricted to small negative penalties on parsing, degeneracy, and per-step tool errors; long-horizon credit assignment is carried entirely by the binary verifier at the end of the trajectory, propagated to every token of the trajectory through the advantage $A_i$.

\subsubsection{Task Mix}
The RL prompt distribution spans three families that share a single tool-use API.

\textbf{Code repository tasks} drawn from a large internal SWE collection as well as a mix of public datasets; rewards come from each task's own per-repository unit tests, executed in an ephemeral sandbox after applying the agent's patch.

\textbf{Shell usage tasks} executed inside ephemeral container sandboxes covering shell scripting, system administration, and long-horizon command-line workflows; rewards come from task-specific shell assertions evaluated against the final container state.

\textbf{Tool-integrated math reasoning}: math problems in which the agent must drive a code-execution tool to compute and verify its final answer, scored by exact numeric-answer match against the gold solution. These trajectories anchor the model's reasoning capability under the same tool-use API as the coding tasks, and prevent regression of the reasoning behavior established in mid-training.

We partition tasks into pass-rate buckets using prior attempts from the initial checkpoint. We drop tasks that were always solved or never solved, and sample the remaining tasks proportional to their $(1 - \mathrm{pass\_rate})$. This allows us to get a distribution skewed to harder but still solvable tasks that preserved a training signal over the multiple epochs.

\begin{notebox}[Model Factory Component: Atlas]\phantomsection\label{sec:atlas}
Atlas is our multi-accelerator inference library, built on top of vLLM~\citep{kwon2023efficient} and consuming Titan's model definitions directly for bit-accurate reference with the trainer. Atlas serves every downstream consumer: automated evaluations, internal inference, customer-facing production traffic, but also crucially RL rollout generation. Any improvements in Atlas immediately benefit all downstream consumers.
\smallskip
To manage traffic across Atlas deployments we run an Envoy-based~\citep{envoy} proxy paired with a custom orchestrator that can stand up new model deployments in seconds. The orchestrator owns session management and routing across the fleet, optimizing KV-cache utilization during agentic reinforcement learning.
\end{notebox}

\subsubsection{Trainer-to-inference Weight Sync}
The trainer and the inference fleet live on physically separate GPU pools but share a high-bandwidth interconnect. Updated weights are sent from trainer GPUs to every inference-replica GPU via our GPU$\leftrightarrow$GPU weight-transfer system~\citep{poolside2025posttraining}. The weight sync uses NCCL point-to-point primitives over GPUDirect RDMA with no intermediate offload to host memory or remote object storage. 
The system handles the asymmetric $n{\to}m$ fan-out between the sharded trainer and the inference fleet of full-weight replicas. In our experiments we kept $m$ between $2n$ and $3n$ depending on model size, which allowed us to strike the right balance between speed and off-policy level.
Transfers run asynchronously, so training continues while weights are in flight.

Broadcasts are issued every $2$ optimizer steps. Two synchronization primitives keep all of the above safe under online RL, irrespective of which \texttt{FP8} mode the inference side is in. Weight broadcasts trigger an inference-side KV-cache reset, so the cache can never mix tokens encoded under different weight versions. Weight updates are configured to block any in-flight rollout step on the update, so no single rollout step straddles a refresh; from each rollout's perspective the policy is piecewise-constant in time, which is what the importance ratio $\rho_t$ in the loss above implicitly assumes.
We also cap the staleness of trajectories used for training at $10$ optimizer steps, which we empirically found to be a good borderline between training speed and stability in our earlier experiments. 
However, with our release ratio between training and inference GPUs we never actually hit this limit, which effectively kept all trajectories non-wasteful.

\subsubsection{FP8 Inference for RL Rollouts}
The trainer keeps master weights and gradient computation in \texttt{BF16}; only the inference path uses \texttt{FP8}, and only on the KV cache for the release runs.

We store the attention KV cache in \texttt{FP8} across the agent's full $131{,}072$-token context window.
Agentic trajectories pack many short assistant turns interleaved with long tool observations; at this context length the KV cache dominates inference-replica memory, and storing it in \texttt{FP8} roughly doubles the number of concurrent trajectories a single replica can carry.
The release-candidate runs for both \lagunaxs and \lagunam used this \texttt{FP8} KV cache only, inference-side weights remained in \texttt{BF16} across the entire RL phase.

During pre-release ablations, we also quantized the inference-side weights to \texttt{FP8} with an in-flight block-wise scheme: weights were re-quantized on the inference replica during each weight sync, with per-block scales computed on the fly without any calibration data.
We observed no measurable hit to stability as gradient norms and reward curves closely tracked the \texttt{BF16}-weight baseline, and the importance ratio $\rho_t$ remained well within the clip bounds. 
The only observable effect was a larger training-inference mismatch measured by per-token KL divergence and absolute log-probability differences. 
However, as it was not obvious how this mismatch would play out over the many-day horizon of an RL run, we decided to keep \texttt{BF16} weights for the release runs.

\section{Quantization}\label{sec:quantization}

Since one of our goals in building \lagunaxs was to enable deployment on low-VRAM devices, the
model's memory footprint was a key consideration. Quantization was therefore a crucial step that we
needed to get right to broaden the model's compatibility with various hardware configurations.

\lagunaxs was pre-trained and fine-tuned using the \texttt{BF16} data type.
We quantized the model's MoE layers to \texttt{FP8}, \texttt{INT4}, and \texttt{NVFP4}. We also
quantized the KV cache to \texttt{FP8}. We primarily used the LLM Compressor framework~\citep{llmcompressor2024}
for quantization due to its tight integration with vLLM~\citep{kwon2023efficient}.

\paragraph{FP8 KV Cache.}
The KV cache was quantized to \texttt{FP8} using \(a_{\max}\)-based scaling with a calibration dataset of
\(128\) long-context agentic trajectories. We use per-tensor scaling for KV cache to maximize
compatibility with various \texttt{FP8} attention implementations.

\paragraph{FP8.}
We found that using SpinQuant~\citep{liu2025spinquant} R1 rotation improved the quality of \texttt{FP8}
quantization without introducing any runtime overhead, and therefore adopted it for our \texttt{FP8}
quantization scheme.

For \texttt{FP8} quantization (\texttt{W8A8}), we used an \(a_{\max}\)-based quantization scheme
with dynamic activation scaling that does not require calibration. Assigning a single scale factor to
each \(128 \times 128\) weight tile and \(1 \times 128\) activation group resulted in no measurable
quality degradation relative to the \texttt{BF16} baseline.

\paragraph{INT4.}
Similarly to our \texttt{FP8} quantization, we used SpinQuant R1 rotation as a pre-processing step
before our \texttt{INT4} quantization.

\begin{figure}[h]
  \centering
  \includegraphics[width=\textwidth]{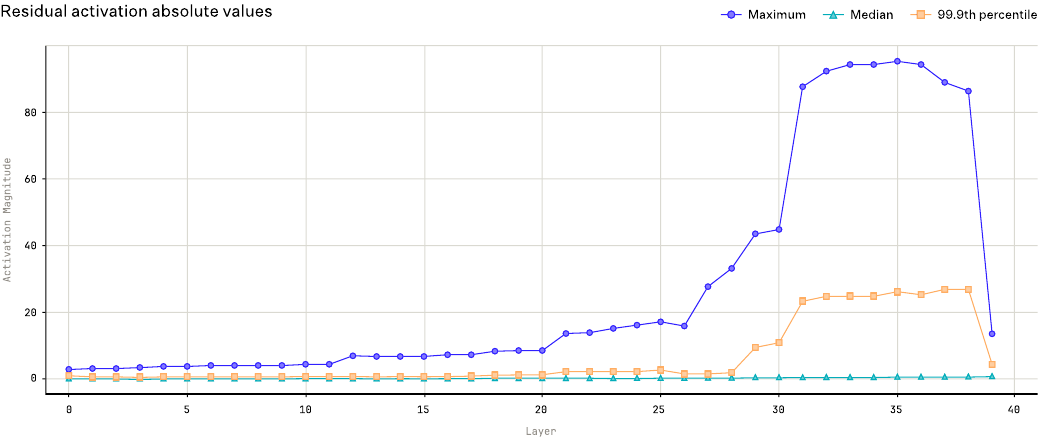}
  \caption{Accumulation of outlier activations across layers. The figure compares absolute maximum
    of the activations in the residual stream to the median of their absolute values.}
  \label{fig:int4_quant_error}
\end{figure}

For \texttt{INT4} weight quantization (\texttt{W4A16}), we applied AWQ~\citep{awq2024} using a
calibration set of 128 long-context agentic trajectories. This approach initially introduced
a non-negligible quality drop. Analysis of the activation distributions in the residual stream of the model
showed an accumulation of outlier activations beginning around layer 30 of the 40-layer network
(see \Cref{fig:int4_quant_error}).

To keep the quality degradation to a minimum, we adopted a mixed-precision quantization strategy: the
first 30 layers were quantized to \texttt{INT4}, while the final 10 layers were quantized to
\texttt{INT8} with \(1 \times 128\) group-based weight scaling.

\paragraph{NVFP4 Quantization.}
Direct post-training \texttt{NVFP4} quantization caused a non-negligible quality drop even in a
mixed-precision \texttt{FP8}/\texttt{NVFP4} strategy mimicking the one we used for \texttt{INT4}.
To mitigate this, we used quantization-aware distillation~\citep{nvidia_nvfp4_qad} (QAD) to recover
quality. Unlike the \texttt{FP8} and \texttt{INT4} schemes above, we did not use SpinQuant for
\texttt{NVFP4} as QAD was sufficient without additional pre-processing. We used the standard
microscaling format for \texttt{NVFP4} following \citet{nvfp4_format}: \(1 \times 16\) groups are
used for local scaling factors in \texttt{FP8} along with an \texttt{FP32} global per-tensor scaling factor.

Starting from the higher-precision checkpoint, we froze all other parameters and optimized the MLP
\texttt{BF16} weights by quantizing/dequantizing to \texttt{NVFP4} in the forward pass. Following
QAD-based accuracy recovery for quantized LLMs~\citep{nvidia_nvfp4_qad}, the quantized student was trained
to match the higher-precision teacher on a fixed distillation dataset. To make the full-vocabulary KL objective
practical, we adopted the hidden-state caching strategy used by DeepSeek-V4~\citep{deepseek_v4}: we cache
final hidden states of the teacher and reconstruct teacher and student logits through the frozen output head.
\paragraph{Quality Evaluation.}
We observed only marginal quality degradation for our final quantization schemes.
However, some of our intermediate quantization attempts resulted in a significant
quality degradation, as described in the paragraphs above. Interestingly, the quality drop of the
unsuccessful attempts was mostly visible on the agentic coding benchmarks, while more traditional
single-turn benchmarks were affected to a smaller extent. This highlights the importance of using
a diverse set of benchmarks to properly validate that the quantization process does not introduce any
unintended side effects, especially for the tasks that require strict formatting such as
agentic coding.

\section{Evaluation}\label{sec:evaluation}
We present evaluation results for \lagunaxs and \lagunam across a variety of benchmarks, including pre-training evaluations for \lagunaxs and agentic evaluations for both \lagunaxs and \lagunam\!\!. 

\begin{notebox}[Model Factory Component: Evaluations]
We schedule configurable evaluations automatically for every training experiment, run against in-flight checkpoints which execute on Atlas (\Cref{sec:atlas}).

A key challenge is \emph{execution-grounded} evaluation: benchmarks where each task requires a real software environment that compiles, runs, and tests the model's output across heterogeneous environments. Here, we use the same code execution component (\Cref{sec:code_exec}) as is used for agentic RL and synthetic data generation.
\end{notebox}

\subsection{Base Model Evaluations}\label{sec:base-evals}

\subsubsection{Evaluation Setup}\label{sec:pretrain-evals-setup}

\paragraph{Benchmarks.} We evaluate \lagunaxs across general, mathematical, and coding capabilities. For general capabilities, we evaluate on BBH~\citep{suzgun2023bbh}, MMLU-STEM~\citep{hendrycks2021mmlu}, and MMLU-Pro~\citep{wang2024mmlupro}. For mathematical reasoning, we use GSM8K~\citep{cobbe2021gsm8k} and MATH~\citep{hendrycks2021math}. For coding capabilities, we employ LiveCodeBench v6~\citep{jain2024livecodebench} (questions from August 2024 to May 2025), MultiPL-E~\citep{cassano2022multiple}, BigCodeBench~\citep{zhuo2024bigcodebench}, and CRUXEval~\citep{gu2024cruxeval}, reported separately with chain-of-thought on input prediction (CRUXEval-I (CoT)) and output prediction (CRUXEval-O (CoT)).

\paragraph{Baselines.} We benchmark \lagunaxs against recent open-weight MoE base models of similar size: Qwen3.5-35B-A3B-Base~\citep{qwen35blog}, Gemma-4-26B-A4B~\citep{gemma4}, and Nemotron-3-Nano-30B-A3B-Base-BF16~\citep{nemotronnano}. We also include MiMo-V2-Flash-Base~\citep{mimo} as a reference point, despite its larger parameter footprint.

\paragraph{Evaluation Configurations.}
We use likelihood-based evaluation for MMLU-STEM and generation-based evaluation for BBH, MMLU-Pro, GSM8K, MATH, LiveCodeBench v6, MultiPL-E, BigCodeBench, CRUXEval-I (CoT), and CRUXEval-O (CoT). Per-benchmark few-shot counts and reported metrics are summarized in \Cref{tab:pretrain-evals}. \lagunaxs scores in \Cref{tab:pretrain-evals} are produced by our internal evaluation framework. For the open-weight baselines, we report published scores when available, which we also reproduce in our internal framework for consistency; specifically, we draw published numbers from the MiMo~\citep{mimo} and Nemotron-3 Nano~\citep{nemotronnano} technical reports. All remaining scores are computed with our internal evaluation framework under the configurations described above.

\subsubsection{Evaluation Results}\label{sec:pretrain-evals-results}

\Cref{tab:pretrain-evals} presents a comparison of \lagunaxs against the four baselines introduced in \Cref{sec:pretrain-evals-setup}.

\begin{table}[!ht]
  \centering
  \footnotesize
  \setlength{\tabcolsep}{5pt}
  \renewcommand{\arraystretch}{1.15}
  \resizebox{\textwidth}{!}{%
  \begin{tabular}{@{}llccccc@{\hspace{1.5em}}c@{}}
    \toprule
    & \textbf{Metric}
    & \textbf{Shots}
    & \shortstack{\textsc{Laguna}\\\textsc{XS.2}}
    & \shortstack{Qwen3.5}
    & \shortstack{Nemotron-3\\Nano}
    & \shortstack{Gemma-4}
    & \shortstack{MiMo-V2\\Flash} \\
    \midrule
    \multicolumn{2}{@{}l}{\# Total Params}    & & 33.4B & 35B & 31.6B & 25.2B & 309B \\
    \multicolumn{2}{@{}l}{\# Active Params}   & & 3B    & 2.6B  & 3.6B  & 3.8B  & 15.1B  \\
    \midrule
    \multicolumn{8}{@{}l}{\textit{General}} \\
    \addlinespace[2pt]
    BBH                   & EM     & 3-shot & 80.9          & \textbf{86.3} & 79.1          & 76.4 & 88.5$^{\dagger}$ \\
    MMLU-STEM             & acc    & 5-shot & 78.1          & \textbf{80.2} & 75.3          & 70.4 & 89.1 \\
    MMLU-Pro              & EM     & 5-shot & 53.0          & 62.5          & \textbf{65.1}$^{\dagger}$ & 47.5 & 73.2$^{\dagger}$ \\
    \midrule
    \multicolumn{8}{@{}l}{\textit{Mathematics}} \\
    \addlinespace[2pt]
    GSM8K                 & EM     & 8-shot & 84.2          & 91.5          & \textbf{92.3}$^{\dagger}$ & 75.4 & 92.3$^{\dagger}$ \\
    MATH                  & EM     & 4-shot & 58.8          & 60.0          & \textbf{82.9}$^{\dagger}$ & 38.9 & 71.0$^{\dagger}$ \\
    \midrule
    \multicolumn{8}{@{}l}{\textit{Coding}} \\
    \addlinespace[2pt]
    LiveCodeBench v6 & pass@1 & 1-shot & \textbf{29.3} & 24.4          & 22.5          & 18.1 & 30.8$^{\dagger}$ \\
    MultiPL-E             & pass@1 & 0-shot & \textbf{58.4}          & 57.9 & 56.1          & 45.1$^{*}$ & 61.1 \\
    BigCodeBench          & pass@1 & 0-shot & \textbf{53.8} & 52.0          & 50.2          & 44.6 & 70.1$^{\dagger}$ \\
    CRUXEval-I (CoT)        & pass@1 & 1-shot & 61.9          & \textbf{66.0} & 63.2          & 54.7 & 67.5$^{\dagger}$ \\
    CRUXEval-O (CoT)        & pass@1 & 1-shot & 71.7 & \textbf{71.9}          & 63.4          & 60.0 & 79.1$^{\dagger}$ \\
    \bottomrule
  \end{tabular}}
  \vspace{5pt}
  \caption{Comparison of \lagunaxs against open-weight base models. EM stands for Exact Match. \textbf{Bold} marks the best score among models of similar size (\lagunaxs\!\!, Qwen3.5-35B-A3B, Nemotron-3-Nano-30B-A3B-BF16, Gemma-4-26B-A4B). $^{\dagger}$Score taken from the corresponding model's technical report (MiMo~\citep{mimo} or Nemotron 3 Nano~\citep{nemotronnano}); all unmarked scores are computed with our internal evaluation framework. $^{*}$For Gemma-4-26B-A4B on MultiPL-E, we use a slightly modified prompt format (a two-space indent appended as a trailing suffix) to work around a prompt-boundary artifact that otherwise significantly degrades performance on \texttt{cpp}, \texttt{sh}, and \texttt{ts}.}
  \label{tab:pretrain-evals}
\end{table}

Our base model results in \Cref{tab:pretrain-evals} place \lagunaxs among comparable open-weight base models. It demonstrates strong performance on coding benchmarks, leading recent open-weight MoE base models of comparable active parameter count on the majority of coding tasks. Moreover, \lagunaxs approaches MiMo-V2-Flash-Base on LiveCodeBench v6 and MultiPL-E despite being a fraction of the size.

While pre-training benchmarks are the most informative signal available at this stage of training, they remain imperfect proxies for downstream task performance and real-world utility. We treat agentic evaluations of the final post-trained model as the primary signal we optimize against. 

\subsection{Agentic Evaluations}

We report performance on the following four benchmarks to best represent the intended use cases of \lagunam and \lagunaxs as agentic coding models: SWE-bench Verified~\citep{jimenez2024swebench, openai2024swebenchverified}, SWE-bench Multilingual~\citep{yang2025swesmith}, SWE-Bench Pro~\citep{scale2025swebenchpro}, and Terminal-Bench 2.0~\citep{tbench2025terminalbench, merrill2026terminalbenchbench}.

\subsubsection{Evaluation Setup}\label{sec:agentic-evals-setup}
\paragraph{Baselines.} Due to the complexity of agentic evaluations, performance can vary drastically depending on configuration, such as harness choice, sandbox resource allocations and sampling parameters. This makes publicly reported results for external models difficult to reproduce with certainty.

To avoid potential bias introduced by our evaluation setup, we report external model baseline scores only from official release blog posts by the authors or equivalent, with the exception of Gemma 4 31B dense, where the highest published scores were reported by the Qwen team~\citep{qwen36_35b_a3b}, and Claude Haiku 4.5, where the highest published (verified) scores for SWE-Bench Pro and Terminal-Bench 2.0 are from their respective official leaderboards.

For \lagunaxs\!\!, we report scores from Devstral Small 2~\citep{mistral2025devstral-2}, Gemma 4 31B dense~\citep{gemma4}, Qwen3.5~\citep{qwen35blog}, Qwen3.6~\citep{qwen36_35b_a3b}, Claude Haiku 4.5~\citep{anthropic2025claudehaiku45}, and GPT-5.4 Nano~\citep{openai2026gpt54nano}.

For \lagunam\!\!, we report scores from Devstral 2~\citep{mistral2025devstral-2}, GLM-4.7~\citep{glm47team2026}, DeepSeek-V4 Flash~\citep{deepseek_v4}, Qwen3.5~\citep{qwen35blog}, and Claude Sonnet 4.6~\citep{anthropic2025claudesonnet46}.

\paragraph{Settings.} All agentic evaluations are run with \emph{pool},\footnote{\url{https://github.com/poolsideai/pool}} our terminal coding agent harness, with a maximum of 500 steps per task. We sample \lagunaxs and \lagunam with a temperature of 1.0 and top\_k = 20, with thinking mode enabled and a context length of 256K tokens. All tasks are run in their own sandbox using our internal sandboxing service. Each sandbox is allocated 2 CPUs and 8 GB of RAM per task with the exception of Terminal-Bench 2.0, for which 32 CPUs and 48 GB of RAM are allocated per sandbox due to more resource-intensive tasks.

Agentic evaluations contain significant noise due to infrastructure limitations inherent to the benchmarks, which can result in large swings in single-trial scores~\citep{anthropic2026benchmarkinfrastructure}. To address this, we run each benchmark four times and report the mean pass@1 across the four runs.

\paragraph{Benchmark Improvements.} Some base task images and verifiers are also patched to fix these infrastructure reliability issues inherent in task setup, such as rate limits on third-party dependencies in external registries used by the verifier. A detailed changelog is provided in \Cref{app:benchmark-fixes}. 

\subsubsection{Benchmark Results}\label{sec:agentic-evals-results}

\begin{table}[!ht]
  \centering
  \footnotesize
  \setlength{\tabcolsep}{5pt}
  \renewcommand{\arraystretch}{1.15}
  \resizebox{\textwidth}{!}{%
  \begin{tabular}{@{}lccccc@{\hspace{1.5em}}cc@{}}
    \toprule
    & \shortstack{\textsc{Laguna}\\\textsc{XS.2}}
    & \shortstack{Devstral\\Small 2}
    & \shortstack{Gemma 4}
    & \shortstack{Qwen3.5}
    & \shortstack{Qwen3.6}
    & \shortstack{Claude\\Haiku 4.5}
    & \shortstack{GPT-5.4\\Nano} \\
    \midrule
    \# Total Params  & 33.4B & 24B  & 31B  & 35B  & 35B  & --   & --   \\
    \# Active Params & 3B    & 24B  & 31B  & 3B   & 3B   & --   & --   \\
    \midrule
    SWE-bench Verified     & 69.9 & 68.0 & 52.0 & 69.2 & \textbf{73.4} & 73.3 & --   \\
    SWE-bench Multilingual & 57.7 & 55.7 & 51.7 & 60.3 & \textbf{67.2} & --   & --   \\
    SWE-Bench Pro          & 46.3 & --   & 35.7 & 44.6 & \textbf{49.5} & 39.5 & 52.4 \\
    Terminal-Bench 2.0     & 35.7 & 22.5 & 42.9 & 40.5 & \textbf{51.5} & 29.8 & 46.3 \\
    \bottomrule
  \end{tabular}}
  \vspace{5pt}
  \caption{Agentic coding results for \lagunaxs against open-weight baselines of comparable size (Devstral Small 2, Gemma 4, Qwen3.5-35B-A3B, Qwen3.6-35B-A3B), with Claude Haiku 4.5 and GPT-5.4 Nano shown as frontier proprietary references of comparable model size. ``--'' indicates a score not reported by the model provider.}
  \label{tab:agentic-evals-xs}
\end{table}

\begin{table}[!ht]
  \centering
  \footnotesize
  \setlength{\tabcolsep}{5pt}
  \renewcommand{\arraystretch}{1.15}
  \resizebox{\textwidth}{!}{%
  \begin{tabular}{@{}lccccc@{\hspace{1.5em}}c@{}}
    \toprule
    & \shortstack{\textsc{Laguna}\\\textsc{M.1}}
    & \shortstack{Devstral 2}
    & \shortstack{GLM-4.7}
    & \shortstack{DeepSeek-V4\\Flash}
    & \shortstack{Qwen3.5}
    & \shortstack{Claude\\Sonnet 4.6} \\
    \midrule
    \# Total Params  & 225B & 123B & 355B & 284B & 397B & --   \\
    \# Active Params & 23B  & 123B & 32B  & 13B  & 17B  & --   \\
    \midrule
    SWE-bench Verified     & 74.6 & 72.2 & 73.8 & \textbf{79.0} & 76.2 & 79.6 \\
    SWE-bench Multilingual & 63.1 & 61.3 & 66.7 & \textbf{73.3} & 69.3 & --   \\
    SWE-Bench Pro          & 49.2 & --   & --   & \textbf{52.6} & 50.9 & --   \\
    Terminal-Bench 2.0     & 45.8 & 32.6 & 41.0 & \textbf{56.9} & 52.5 & 59.1 \\
    \bottomrule
  \end{tabular}}
  \vspace{5pt}
  \caption{Agentic coding results for \lagunam against open-weight MoE and dense baselines (Devstral 2, GLM-4.7, DeepSeek-V4-Flash, Qwen3.5-397B-A17B), with Claude Sonnet 4.6 shown as a frontier proprietary reference of comparable model size. ``--'' indicates a score not reported by the model provider.}
  \label{tab:agentic-evals-m}
\end{table}

\subsubsection{Reward Hacking}
At the time of writing, we have discovered that the official versions for all four agentic benchmarks reported are vulnerable to benchmark hacking to some degree~\citep{poolside2026benchmarkhacking}, either via leaked git history in the task images or via web search for reference solutions. Evidence of benchmark hacking has also been found on public leaderboard results\footnote{\url{https://www.tbench.ai/news/leaderboard-integrity-update}}.

To address this, we patched the base task images to remove the git history leak for all affected benchmarks, and opened issues and pull requests to contribute back these fixes to the official benchmark repositories. All scores in this report are run with the patched images.

Additionally, we have developed a reward hack judge to detect these specific cheating strategies, which we ran post-hoc on all \lagunam and \lagunaxs evaluation runs for this report. After joint judge review and manual review, we did not find significant reward hacking in the runs. 

\section{Conclusion}

We presented \lagunam and \lagunaxs\!\!, two Mixture-of-Experts foundation models for long-horizon, agentic coding. Both models are competitive with state-of-the-art open models in their respective weight classes on agentic software engineering and terminal benchmarks (\Cref{sec:evaluation}).

We believe in the merits of open research and have shared deep technical details about the research that went into our models. In addition to this, we have also released the weights of \lagunaxs to the public under the Apache~2.0 license. We hope this contributes to the growth of a vibrant and diverse research community in AI.

Moreover, we described the principles of our \emph{Model Factory} that accelerates our research and model building progress. While the architectural, data, and post-training choices that shaped \lagunam and \lagunaxsshort are interesting in their own right, we also want to draw attention to the process behind them.

We believe this is one of the most consequential lever for frontier model development going forward. As models, data pipelines, and post-training recipes grow in complexity, the gap between teams that treat model building as a craft and those that treat it as an industrial process will continue to widen. Investing in the factory is how we expect to keep iteration speed ahead of model complexity.

\lagunam and \lagunaxs are early outputs of this approach. We are continuing to build the factory, and look forward to sharing future models and our research.

\section{Contribution}
Authors are listed alphabetically by their last name. Names with * have since left Poolside.

\paragraph{Research \& Engineering.} Julien Abadji, Marah Abdin, Connor Adams, Eric Alcaide, Mustafa Altun, Michele Artoni, Junze Bao, Uday Barar, Vassilis Bekiaris, Arkadii Bessonov, Benjamin Bütikofer, Jonathan Chang, Yen-Chun Chen, Dmitry Chernenkov, Yang Chi, Filippos Christianos, Fenia Christopoulou, Razvan-Andrei Ciocoiu, Tzachi Cohen, Yohann Coppel, Dmitrii Emelianenko, Brandon Fergerson, Brian Fitzgerald, Matthias Gallé, Alex Golonzovskyi, George Grigorev, Yiyang Hao*, Christian Hensel, Jan Huenermann*, Ye Ji, Sarthak Joshi*, Eiso Kant, Kabir Khandpur, Seonghyeon Kim, Vladimir Kirichenko, Umut Kocasarac, Ilya Kochik, Ivan Komarov, Chaerin Kong, Anurag Koul*, François-Joseph Lacroix, Sergei Laktionov, Waren Long, Quentin Malartic*, Vadim Markovtsev, Afonso Marques, Robert McHardy, Carlos Mocholí, Dmitry Monakhov, Adam Morris, Martin Muller, Christian Mürtz, Robin Nabel, Thien Nguyen*, Rok Novosel, Szymon Ozog, Aalhad Patankar, Aleksei Petrov, Alexandre Piché*, Arthur Pignet, Teodor Poncu, Phil Potter, Alexander Rakowski, Pierre-Yves Ritschard, Jay Roberts, Joe Rowell, Piotr Sarna, Pierre-André Savalle*, Uladzislau Sazanovich, Nikita Shapovalov, Arsenii Shevchenko, Mikhail Shilkov, Andrei Sokol, Mohamed Soliman, Jack Stephenson, Victor Storchan, Dragos-Constantin Tantaru, Artem Tyurin, Adrian Wälchli, Pengming Wang, Jianxiao Yang, Renat Zayashnikov, Alexander Zelenka Martin, Nikolay Zinov.

\paragraph{Partnerships \& Delivery.} Caroline Bercier, José Caldeira, Margarida Garcia, Tom George, Kabeer Gharzai, Glenn Hitchcock, Carson Klingenberg, Ivo Pinto, Varun Randery, Noah Smith, Arina Sugako, Jason Warner.

\printbibliography

\appendix
\crefalias{section}{appendix}
\crefalias{subsection}{appendix}
\section{Additional Details}

\subsection{WSD Optimal-LR Scaling Law}\label{app:lr_scaling}

Supporting figures and tables for the WSD optimal-LR scaling law described in \Cref{sec:pretraining_lr_law}. All sweep runs use four model sizes (2B, 4B, 8B, 16B total parameters), six learning rates from $8{\times}10^{-5}$ to $6{\times}10^{-3}$, fixed global batch $B_0 = 8.4$M tokens, and the stage-1 pre-training mix.

\begin{figure}[!htbp]
  \centering
  \includegraphics[width=0.95\textwidth]{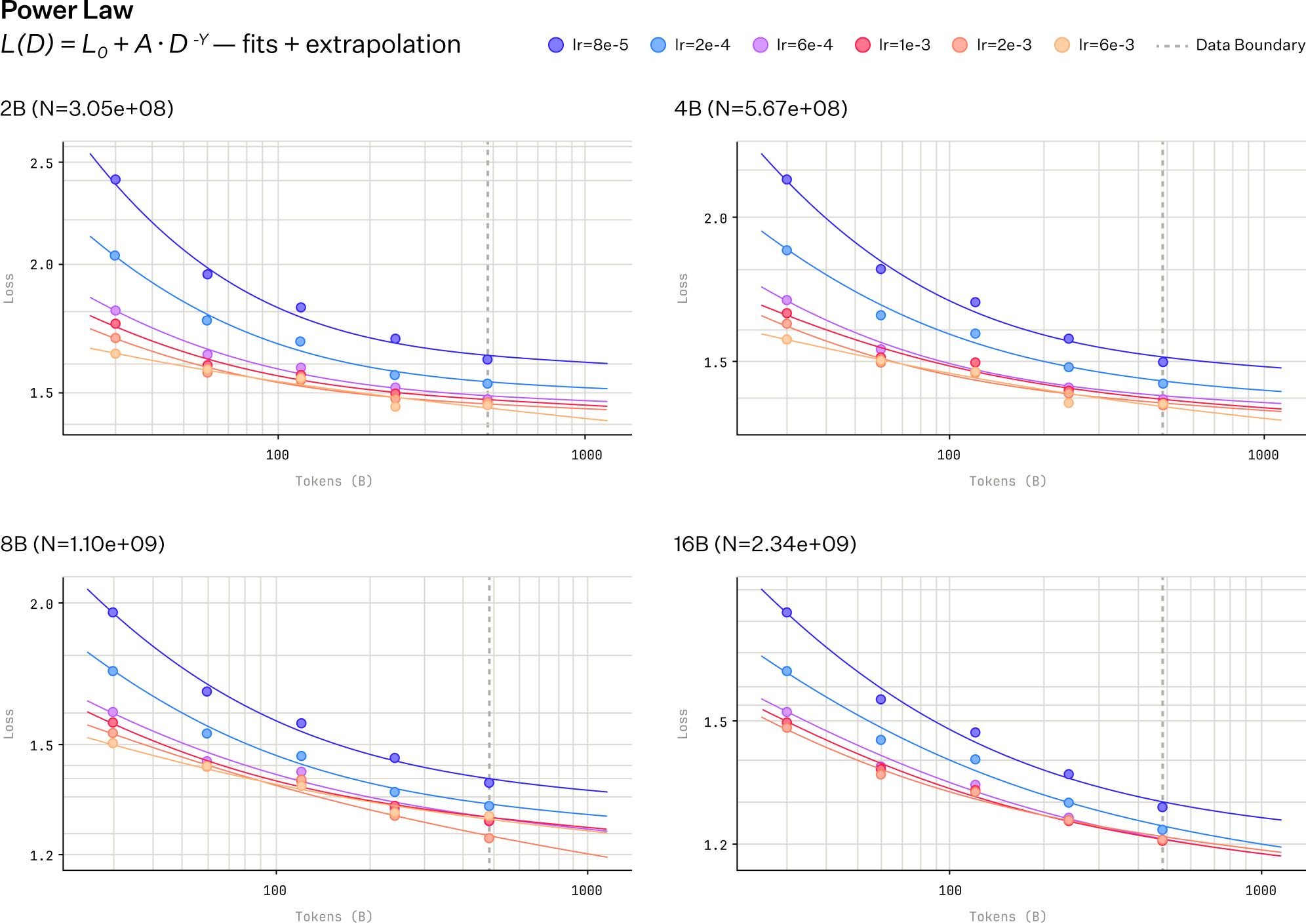}
  \caption{Loss-vs-tokens power-law fits $L(D) = L_0 + A\,D^{-\gamma}$ for each model size, one curve per learning rate. Dashed vertical lines mark the 480B-token data boundary; curves are extrapolated to $\sim$960B tokens.}
  \label{fig:lr_powerlaw_grid}
\end{figure}

\begin{figure}[!htbp]
  \centering
  \includegraphics[width=0.7\textwidth]{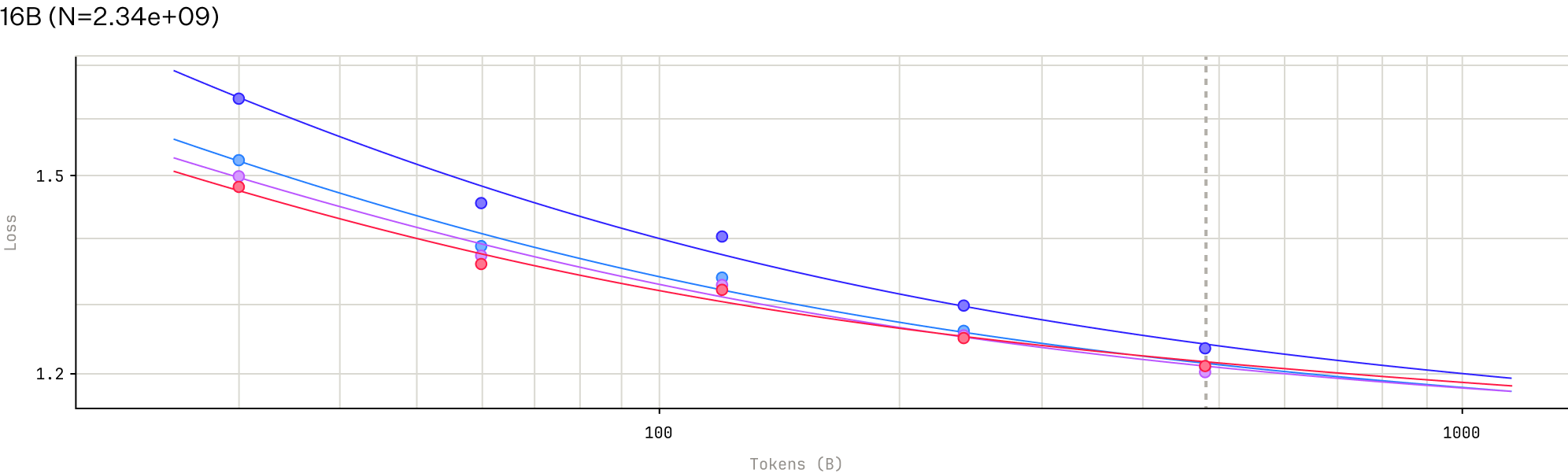}
  \caption{Same fit as \Cref{fig:lr_powerlaw_grid} for the 16B model with the lowest learning rate ($8{\times}10^{-5}$) excluded.}
  \label{fig:lr_powerlaw_16b}
\end{figure}

\paragraph{Per-cell fits.} Per-$(N, \text{lr})$ power-law fits $L(D) = L_0 + A\,D^{-\gamma}$ across the five cooldown points are shown in \Cref{fig:lr_powerlaw_grid}; \Cref{fig:lr_powerlaw_16b} zooms in on the 16B model.

\begin{figure}[!htbp]
  \centering
  \includegraphics[width=0.95\textwidth]{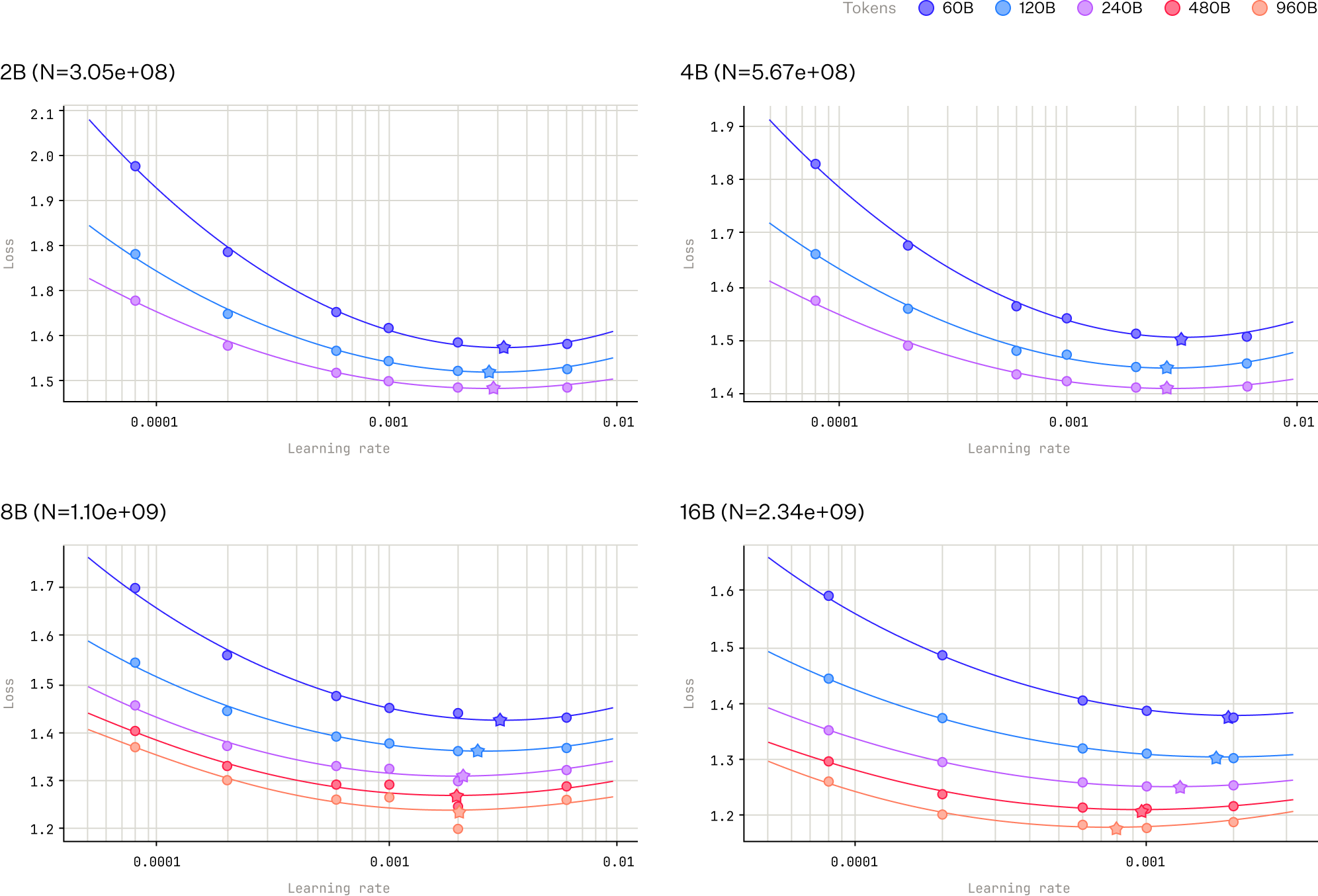}
  \caption{Per-$(N, D)$ parabola fits with vertex $\text{lr}^\star$ marked by stars.}
  \label{fig:lr_parabolas}
\end{figure}

For each $(N, D \in \{60, 120, 240, 480, 960\}\text{B})$ we fit a parabola in $\log_{10}(\text{lr})$ space and take the vertex as $\text{lr}^\star(N, D)$ (\Cref{fig:lr_parabolas}); rows with curvature $a < 0.065$ (flat parabolas, optimum poorly constrained) are excluded from the global fit.

\paragraph{Fit robustness.} A companion fit using a stricter curvature threshold ($a \geq 0.065$, 16 retained points) yields $(\ell_0, \alpha, \beta) = (4.292, -0.4393, -0.2689)$ with $R^2 = 0.8865$ and per-size mean residuals (in $\log_{10}\text{lr}$) of $+0.001 / -0.041 / +0.058 / -0.022$ for the 2B / 4B / 8B / 16B groups (multiplicative LR errors $1.00\times / 0.91\times / 1.14\times / 0.95\times$); the headline form in \Cref{eq:lr_law} agrees within the LR-extrapolation uncertainty for \lagunaxs\!\!.

\paragraph{External cross-check.} Applied to Kimi K2~\citep{kimik2}
($N = 32.6$B active, $D = 15.5$T tokens, $B = 67$M), the law predicts
$\sim 3.5{\times}10^{-4}$; Kimi K2 was trained at $2{\times}10^{-4}$.
The two differ within an order of magnitude despite Kimi K2 using a
$35\%$ cooldown (vs.\ our $30\%$), a different MoE sparsity ratio,
MuonClip, aux-loss-free load balancing, and different data, so this
should be read as suggestive rather than a validation of the law
outside our regime.

\FloatBarrier

\subsection{SWA Architecture Ablations}\label{app:swa_ablations}

Supporting tables for the SWA architecture summary in \Cref{sec:architecture}. All ablation rows use a 16B-total / 2.3B-active MoE model; hyperparameters are included in \Cref{tab:swa_ablation_hparams}. Only the attention design (window, gating, $\theta_{\text{swa}}$, RoPE coverage, head allocation) varies between rows.

The 4K Avg is calculated over 10 base benchmarks (PiQA~\citep{bisk2020piqa}, WinoGrande~\citep{sakaguchi2019winogrande}, ARC-Easy \& ARC-Challenge~\citep{clark2018thinksolvedquestionanswering}, HellaSwag~\citep{zellers-etal-2019-hellaswag}, APTBench-4k~\citep{qin2025aptbenchbenchmarkingagenticpotential}, MMLU~\citep{hendrycks2021mmlu}, GPQA-Diamond~\citep{rein2024gpqa}, MultiPL-E~\citep{cassano2022multiple}, EvalPlus~\citep{liu2023is}); the 32K and 128K Avgs are each calculated over 4 tasks at the corresponding window (APTBench, RULER QA, GSMInfinite, LongBenchV2-Academic).

Five reruns of the baseline give a standard deviation of $\sim$0.0026 on the 4K Avg, so we treat differences below $\sim$0.005 as within noise.

\begin{table}[H]
\centering
\small
\begin{tabular}{lc}
\toprule
\textbf{Hyperparameter} & \textbf{Value} \\
\midrule
Total / active parameters                 & 16.6B / 2.3B \\
Layers                                    & 22 \\
Hidden size $d_{\text{model}}$            & 2048 \\
FFN intermediate (dense)                  & 8192 \\
FFN intermediate (per routed expert)      & 1024 \\
Routed experts (top-$k$)                  & 128 (top-8) \\
Shared experts                            & 1 \\
Optimizer                                 & Muon \\
LR scheduler                              & Cosine \\
\midrule
\multicolumn{2}{l}{\emph{Pre-training}} \\
Sequence length                           & 4096 \\
Global batch size                         & 8.4M tokens \\
Total tokens / steps                      & 335B / 40{,}000 \\
Peak LR                                   & $2{\times}10^{-3}$ \\
Warmup steps                              & 1{,}000 \\
\midrule
\multicolumn{2}{l}{\emph{32K / 128K continued pre-training (long-context)}} \\
Sequence length                           & 32{,}768 / 131{,}072 \\
Global batch size                         & 8.4M tokens \\
Total tokens / steps                      & 50B / 6{,}000 \\
Peak LR (32K / 128K)                      & $7{\times}10^{-5}$ / $7{\times}10^{-6}$ \\
Warmup steps                              & 200 \\
\bottomrule
\end{tabular}
\caption{Ablation model hyperparameters.}
\label{tab:swa_ablation_hparams}
\end{table}

\begin{table}[H]
\centering
\small
\begin{tabular}{lccc}
\toprule
\textbf{Architecture} & \textbf{4K Avg $\uparrow$} & \textbf{32K Avg $\uparrow$} & \textbf{128K Avg $\uparrow$} \\
\midrule
Dense GA, full RoPE, full gating                                & 0.5389 & \textbf{0.308} & 0.290 \\
\;\;+ SWA-1024 (interleaved 3:1)                                & 0.5292 & 0.274 & 0.267 \\
\;\;+ per-head gating, $\theta_{\text{swa}}=10{,}000$           & 0.5328 & 0.267 & 0.272 \\
\;\;+ GA Partial RoPE (50\%)                                    & 0.5425 & 0.266 & 0.284 \\
\;\;+ SWA-512                                                   & 0.5449 & 0.285 & \textbf{0.304} \\
\;\;+ 48 GA / 64 SWA Q-heads, $k_{\text{dense}}{=}1$ (final ablation architecture) & \textbf{0.5455} & 0.305 & 0.296 \\
\bottomrule
\end{tabular}
\caption{Architecture ablations across short and long contexts.}
\label{tab:swa_ablations}
\end{table}
\bigskip

\subsection{Benchmark Improvements}\label{app:benchmark-fixes}

Many benchmarks contain components which encounter reliability issues when run at scale, such as rate limits on third-party dependencies, unpinned dependency versions, and behavior unique to our sandbox. To combat infrastructure noise outside of our control, we made numerous improvements to the agentic benchmarks reported on.

\paragraph{Terminal-Bench 2.0.}

Many of the tasks in Terminal-Bench 2.0 rely on third-party dependencies hosted on external registries, which have rate limits that can be easily hit when running the benchmark at scale. Just like many other benchmarks, external dependencies can also be pulled or become stale, so we also updated numerous tasks to rely on updated dependency versions.

\begin{itemize}
  \item All tasks were run with a 3-hour timeout, 32 CPUs, 48 GB RAM, and 50 GB storage.
  \item We added in-task retries on components that rely on external services which are known to be flaky
  \item We vendored (stored copies in internal storage) all external dependencies for tasks \texttt{fix-ocaml-gc}, \texttt{reshard-c4-data}, \texttt{build-cython-ext}, \texttt{filter-js-from-html}, \texttt{sam-cell-seg}, \texttt{pytorch-model-cli}, \texttt{install-windows-3.11} and bundle the version of \texttt{uv}/\texttt{uvx} required by all images. This greatly reduced failures due to external rate limits.
  \item Tasks \texttt{qemu-alpine-ssh} and \texttt{qemu-startup} were updated to resolve a problem where open telnet sessions could not be opened in the verification stage, resulting in failures due to inability to grade results. 
  \item Task \texttt{polyglot-c-py} replaced the strict assertion that \texttt{main.py.c} is the only file in the directory with an assertion that the file exists in the directory. Without this change, a valid solution (compiling in-place to produce a binary alongside \texttt{main.py.c}) could fail due to test-time requirements that contradict the instructions which allow this solution pattern.
  \item Fixed dependency drift on tasks: \texttt{build-pmars}, \texttt{crack-7z-hash}, \texttt{mteb-leaderboard}, \texttt{mteb-retrieve}, \texttt{rstan-to-pystan}, \texttt{dna-assembly}, \texttt{dna-insert}, \texttt{hf-model-inference}, and \texttt{kv-store-grpc}. Dependency drift occurs when a task's dependencies change from what the original build expected, resulting in broken builds or tests in the environment.
\end{itemize}

We are collaborating with the Harbor team to merge these changes upstream where possible.

\paragraph{SWE-bench Verified.}

The upstream SWE-bench Verified dataset is not frequently updated; running with the canonical dataset results in dependency and fixture drift. Our setup also faced rate limit failures from external providers, much like Terminal-Bench 2.0, due to the parallelism at which we run evaluation. 

\begin{itemize}

  \item The image for task \texttt{astropy-8872} is pinned to \texttt{setuptools==68.0.0} which raises a deprecated version warning and errantly fails the grader in Harbor. To fix this we pinned to \texttt{setuptools==58.0.0}.
  \item In Task \texttt{astropy-7606}, pytest emits \texttt{test\_compose\_roundtrip[unit0]} but the dataset expects \texttt{test\_compose\_roundtrip[]}. We fixed this mismatch in our setup.
  \item Task \texttt{sphinx-8475} has tests which rely on a \texttt{w3.org} URL which returns a 403. The task was fixed to allow for this instead of errantly failing. 
  \item The pass-to-pass tests for Task \texttt{django-10097} fail with the golden patch due to Django/SQLite teardown issues unrelated to the task. If the known teardown flake occurs with the exact signature seen under the golden patch, and is the only failure, we ignore it. 
  \item Eight of the \texttt{psf/requests} repo tasks call a \texttt{httpbin()} helper which defaults to a \texttt{httpbin.org} endpoint which under heavy load can return a transient 5xx error. We run a sidecar for HTTP and HTTPS requests to handle these requests instead with a fallback to the default \texttt{httpbin.org} endpoint.

\end{itemize}

\paragraph{SWE-bench Multilingual.}

Our SWE-bench Multilingual changes focused on improving the reliability of the dataset when run at scale with high parallelism.

\begin{itemize}
  \item All images were updated to have future git tags removed to prevent possible reward hacking.
  \item Task \texttt{apache\_\_lucene-12626} was updated to have additional network retries due to higher than normal network flakes. 
  \item Task \texttt{tokio-rs\_\_tokio-4384} fails due to dependency mismatches; we pin dependencies (\texttt{rand 0.8.5}, \texttt{rand\_core 0.6.4}, and \texttt{getrandom 0.2.15}).
  \item Task \texttt{tokio-rs\_\_tokio-6838} deterministically fails due to pass-to-pass test \texttt{uds\_stream::epollhup} returning \texttt{Ok(\_)} where the test expects \texttt{Err(ConnectionReset | ConnectionRefused)}. This issue was due to our sandbox kernel's epoll readiness not including the expected \texttt{EPOLLHUP} bit when a Unix-socket listener is dropped mid-connect. This resulted in an \texttt{Ok(\_)} signal and subsequent unit test failures in the repo's default state. The task was updated to allow \texttt{Ok(\_)} signals in the pass-to-pass check.
  \item Task \texttt{micropython\_\_micropython-12158} deadlocks in our sandbox service due to thread starvation; we fixed this by updating the \texttt{tests/thread/thread\_exc1.py} test to use \texttt{time.sleep(0)} instead of \texttt{pass}. 

\end{itemize}

\paragraph{SWE-Bench Pro.}

For SWE-Bench Pro we made edits focused on fixing verifier selections and the previously mentioned reward hack vulnerabilities.

\begin{itemize}
  \item All images were updated to have git history after the checked-out commit fully removed to prevent potential reward hacking. See \url{https://github.com/harbor-framework/harbor/pull/1593}
  \item Tasks \texttt{future-architect\_\_vuls} and \texttt{ansible\_\_ansible} had incorrect fail-to-pass or pass-to-pass names due to trailing punctuation.
  \item Task \texttt{tutao\_\_tutanota} contained incorrect suffixes for its subset aggregations; this was fixed by dropping the \texttt{(\textbackslash{}d+ assertions)} prior to comparison.

\end{itemize}

\end{document}